\documentclass{pnastwo}

\usepackage{pnastwoF}
\usepackage{amssymb,amsfonts,amsmath}
\usepackage{cite}

\usepackage{textcomp}

\usepackage{graphicx}
\usepackage{sidecap}
\graphicspath{{./figs/}}
\usepackage{epstopdf}
\usepackage{ifthen}
\usepackage{color}
\newcommand{\beq}[1][]{\begin{equation} \ifthenelse{\equal{#1}{}}{}{\label{#1}}}
\newcommand{\eeq}{\end{equation} }
\newcommand{\beqr}{\begin{eqnarray}}
\newcommand{\eeqr}{\end{eqnarray} }

 \newcommand{\req}[1]{(\ref{#1})}

\newcommand{\wa}{{{\bf W}^{1}}}
\newcommand{\wb}{{ {\bf W}^{2}}}
\newcommand{\ws}{{\bf W}^s}
\newcommand{\sio}{{\bf \Sigma}^{yx}}
\newcommand{\si}{{\bf \Sigma}^{x}}
\newcommand{\rs}{{\bf V}}
\newcommand{\ls}{{\bf U}}
\newcommand{\sv}{{\bf S}}

\newcommand{\at}{{{\bf A}(t)}}

\newcommand{\br}{{\bf R}}

\newcommand{\ddt}{\frac{d}{dt}}

\newcommand{\x}{{\bf x}}
\newcommand{\y}{{\bf y}}
\newcommand{\h}{{\bf h}}
\newcommand{\uv}{{\bf u}}
\newcommand{\vv}{{\bf v}}
\newcommand{\ov}{{\bf o}}

\newcommand{\sh}{{\bf \Sigma}^{h}}
\newcommand{\shy}{{\bf \Sigma}^{\hat y}}

\newcommand{\figsetup}{Fig.\,1}
\newcommand{\figprogdiff}{Fig.\,2}
\newcommand{\figsvd}{Fig.\,3}
\newcommand{\fighiersvd}{Fig.\,4}
\newcommand{\figilluscorrs}{Fig.\,5}
\newcommand{\figtypgeom}{Fig.\,6}
\newcommand{\figcategcoh}{Fig.\,7}
\newcommand{\figbasiclvl}{Fig.\,8}
\newcommand{\figstructform}{Fig.~9}
\newcommand{\figinductgen}{Fig.\,10}
\newcommand{\figlearninginvar}{Fig.\,11}

\issuedate{}
\volume{}
\issuenumber{}
\begin{document}

\title{A mathematical theory of semantic development in deep neural networks}

\author{Andrew M. Saxe\affil{1}{University of Oxford, Oxford, UK},
 	James L. McClelland\affil{2}{Stanford University, Stanford, CA},    
	\and
	Surya Ganguli\affil{2}{}\affil{3}{Google Brain, Mountain View, CA}}

\contributor{}

\maketitle

\begin{article}
\begin{abstract}

An extensive body of empirical research has revealed remarkable regularities in the acquisition, organization, deployment, and neural representation of human semantic knowledge, thereby raising a fundamental conceptual question: what are the theoretical principles governing the ability of neural networks to acquire, organize, and deploy abstract knowledge by integrating across many individual experiences?  We address this question by mathematically analyzing the nonlinear dynamics of learning in deep linear networks.  We find exact solutions to this learning dynamics that yield a conceptual explanation for the prevalence of many disparate phenomena in semantic cognition, including the hierarchical differentiation of concepts through rapid developmental transitions, the ubiquity of semantic illusions between such transitions, the emergence of item typicality and category coherence as factors controlling the speed of semantic processing, changing patterns of inductive projection over development, and the conservation of semantic similarity in neural representations across species.  Thus, surprisingly, our simple neural model qualitatively recapitulates many diverse regularities underlying semantic development, while providing analytic insight into how the statistical structure of an environment can interact with nonlinear deep learning dynamics to give rise to these regularities.

\end{abstract}


\abbreviations{SVD, singular value decomposition}

\dropcap{H}uman cognition relies on a rich reservoir of semantic knowledge enabling us to organize and reason about our complex sensory world \cite{Keil1979, Carey1985, Murphy2002, Rogers2004}.  This semantic knowledge allows us to answer basic questions from memory (i.e. "Do birds have feathers?"), and relies fundamentally on neural mechanisms that can organize individual items, or entities (i.e. \textit{Canary}, \textit{Robin}) into higher order conceptual categories (i.e. \textit{Birds}) that include items with similar features, or properties. This knowledge of individual entities and their conceptual groupings into categories or other ontologies is not present in infancy, but develops during childhood \cite{Keil1979, Mandler1993}, and in adults, it powerfully guides the deployment of appropriate inductive generalizations.

The acquisition, organization, deployment, and neural representation of semantic knowledge has been intensively studied, yielding many well-documented empirical phenomena.  For example, during acquisition, broader categorical distinctions are generally learned before finer-grained distinctions \cite{Keil1979, Mandler1993}, and long periods of relative stasis can be followed by abrupt conceptual reorganization \cite{Inhelder1958,Siegler1976}. Intriguingly, during these periods of developmental stasis, children can strongly believe illusory, incorrect facts about the world \cite{Carey1985}.  

Also, many psychophysical studies of performance in semantic tasks have revealed empirical regularities governing the organization of semantic knowledge.   In particular, category membership is a {\it graded} quantity, with some items being more or less typical members of a category (i.e. a sparrow is a more typical bird than a penguin).  The notion of item typicality is both highly reproducible across individuals \cite{Rosch1975,Barsalou1985} and correlates with performance on a diversity of semantic tasks \cite{Rips1973,Murphy1985a,Mervis1976,Rips1975,Osherson1990}.  Moreover, certain categories themselves are thought to be highly coherent (i.e. the set of things that are \textit{Dogs}), in contrast to less coherent categories (i.e. the set of things that are \textit{Blue}).  More coherent categories play a privileged role in the organization of our semantic knowledge; coherent categories are the ones that are most easily learned and represented \cite{Rosch1975,Rosch1978,Murphy1985}.  Also, the organization of semantic knowledge powerfully guides its deployment in novel situations, where one must make inductive generalizations about novel items and properties  \cite{Carey1985,Murphy2002}.  Indeed, studies of children reveal that their inductive generalizations systematically change over development, often becoming more specific with age \cite{Carey1985, Murphy2002, Keil1991, Carey2011,Gelman1990a}.

Finally, recent neuroscientific studies have begun to shed light on the organization of semantic knowledge in the brain.  The method of representational similarity analysis  \cite{Edelman1998,Kriegeskorte2013} has revealed that the similarity structure of neural population activity patterns in high level cortical areas often reflect the semantic similarity structure of stimuli, for instance by differentiating inanimate objects from animate ones \cite{Carlson2014,Carlson2013,Giordano2013,Liu2013,Connolly2012}. And strikingly, studies have found that such neural similarity structure is preserved across human subjects, and even between humans and monkeys \cite{Mur2013,Esteky2011}.

This wealth of empirical phenomena raises a fundamental conceptual question about how neural circuits, upon experiencing many individual encounters with specific items, can over developmental time scales extract abstract semantic knowledge consisting of useful categories that can then guide our ability to reason about the world and inductively generalize.  While a diverse set of theories have been advanced to explain human semantic development, there is currently no analytic, mathematical theory of neural circuits that can account for the diverse phenomena described above.  Interesting non-neural accounts for the discovery of abstract semantic structure include for example the conceptual "theory-theory"  \cite{Carey1985,Keil1991,Murphy1985,Carey2011}, and computational Bayesian \cite{Kemp2008} approaches.  However, neither currently proposes a neural implementation that can infer abstract concepts from a stream of specific examples. %"Theory-theory" conceptualizes the existence of theories of the learning agent that can abruptly change, but this compelling conceptual account leaves open the computational details underlying theory change.   Bayesian inference methods \cite{Kemp2008}, on the other hand, provide a computationally well-defined and transparent method for extracting the best semantic structure that explains a dataset, but currently awaits a neurally plausible instantiation that can infer latent concepts from specific examples as they are received. % Indeed, a naive application of the Bayesian approach requires storing the entire history of specific examples and rerunning a complicated inference engine upon receipt of each new example.
In contrast, much prior work has shown, through simulations, that neural networks can gradually extract semantic structure by incrementally adjusting synaptic weights via error-corrective learning \cite{Hinton1986-cw,Rogers2004,Rumelhart1993,McClelland2005,Plunkett1992,Quinn1997,Colunga2005,McMurray2012,Blouw2016}. However, in contrast to the computational transparency enjoyed by Bayesian approaches, the theoretical principles governing how even simple artificial neural networks extract semantic knowledge from their ongoing stream of experience, embed this knowledge in their synaptic weights, and use these weights to perform inductive generalization, remains obscure.  
 
 In this work, our fundamental goal is to fill this gap by employing an exceedingly simple class of neural networks, namely deep linear networks.  Surprisingly, we find that this model class can qualitatively account for a diversity of phenomena involving semantic cognition described above. Indeed, we build upon a considerable neural network literature \cite{Hinton1986-cw,Rumelhart1993,McClelland2005,Plunkett1992,Quinn1997,Colunga2005,McMurray2012,Blouw2016} addressing such phenomena through simulations of more complex nonlinear networks. We build particularly on the integrative, simulation-based treatment of semantic cognition in \cite{Rogers2004}, often using the same modeling strategy in a simpler linear setting, to obtain similar results but with additional analytical insight.  Thus, in contrast to prior work, whether conceptual, Bayesian, or connectionist, our simple model is the first to permit exact analytical solutions describing the entire developmental trajectory of knowledge acquisition and organization, and its subsequent impact on the deployment and neural representation of semantic structure.  In the following, we describe each of these aspects of semantic knowledge acquisition, organization, deployment, and neural representation in sequence, and we summarize our main findings in the discussion. 
   
\section{A Deep Linear Neural Network Model}

Here we consider a framework for analyzing how neural networks extract abstract semantic knowledge by integrating across many individual experiences of items and their properties, across developmental time.  In each experience, given an item as input, the network is trained to correctly produce its associated properties or features as output. Consider for example, the network's interaction with the semantic domain of living things, schematized in \figsetup A.  If the network encounters an item, such as a \textit{Canary}, perceptual neural circuits produce an activity vector $\x \in \bf R^{N_1}$ that identifies this item and serves as input to the semantic system.  Simultaneously,  the network observes some of the item's properties, for example that a canary \textit{Can Fly}. Neural circuits produce an activity feature vector $\y \in \bf R^{N_3}$ of that item's properties which serves as the desired output of the semantic network.  Over time, the network experiences many individual episodes with a variety of different items and their properties. The total collected experience furnished by the environment to the semantic system is thus a set of $P$ examples $\left\{\x^i,\y^i\right\}, i=1,\ldots, P$, where the input vector $\x^i$ identifies item $i$, and the output vector $\y^i$ is a set of features to be associated to this item.

\begin{figure}
\begin{center}
\includegraphics[width=3.5in]{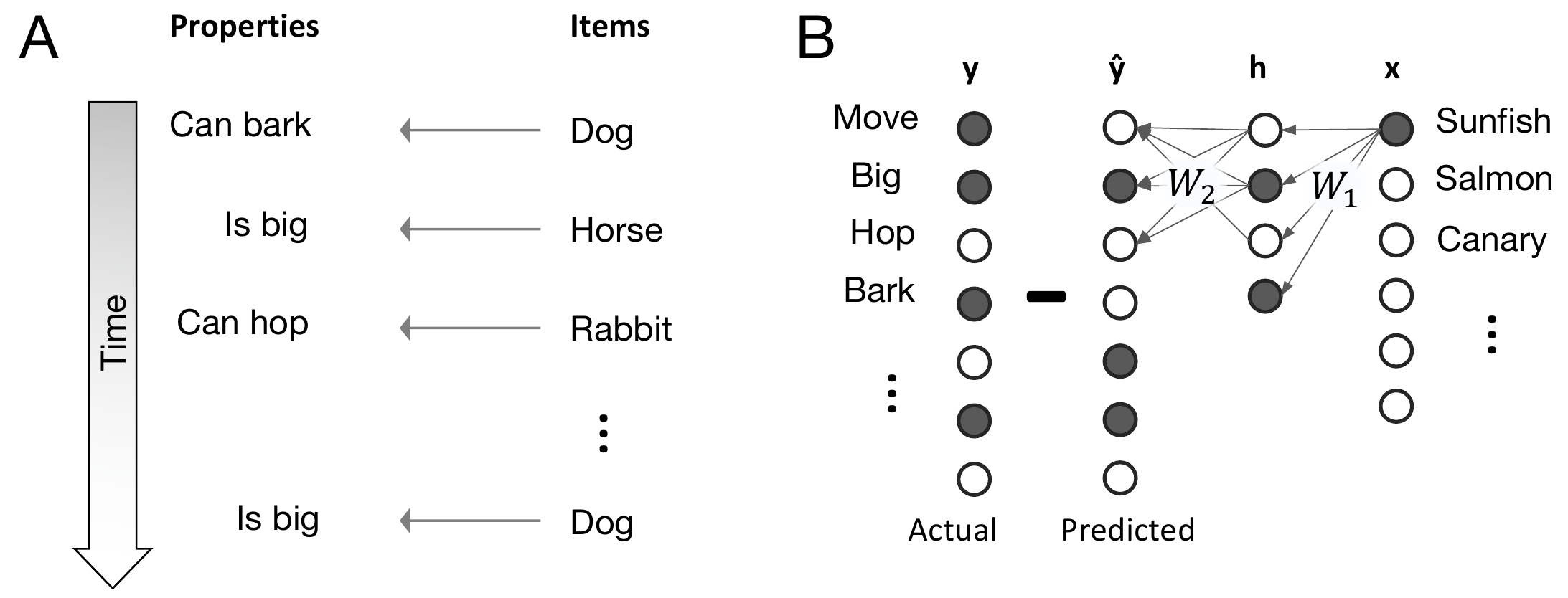}
\end{center}
\caption{(A) During development, the network experiences sequential episodes with items and their properties. (B) After each episode, the network adjusts its synaptic weights to reduce the discrepancy between actual observed properties $\y$ and predicted properties $\hat \y$.  
}
\label{fig1}
\end{figure}

The network's task is to predict an item's properties $\y$ from its perceptual representation $\x$. These predictions are generated by propagating activity through a three layer linear neural network (\figsetup B). The input activity pattern $\x$ in the first layer propagates through a synaptic weight matrix $\wa$ of size $N_2 \times N_1$, to create an activity pattern $\h=\wa \x$ in the second layer of $N_2$ neurons. We call this layer the ``hidden''  layer because it corresponds neither to input nor output.  The hidden layer activity then propagates to the third layer through a second synaptic weight matrix $\wb$ of size $N_3 \times N_2$, producing an activity vector $\hat \y = \wb \h$ which constitutes the output prediction of the network. The composite function from input to output is thus simply $\hat \y = \wb \wa \x$. For each input $\x$, the network compares its predicted output $\hat \y$ to the observed features $\y$ and it adjusts its weights so as to reduce the discrepancy between $\y$ and $\hat \y$.  

To study the impact of depth, we will contrast the learning dynamics of this deep linear network to that of a shallow network that has just a single weight matrix, $\ws$ of size $N_3 \times N_1$ linking input activities directly to the network's predictions $\hat \y = \ws \x$. At first inspection, it may seem that there is no utility whatsoever in considering deep linear networks, since the composition of linear functions remains linear.  Indeed, the appeal of deep networks is thought to lie in the increasingly expressive functions they can represent by successively cascading many layers of nonlinear elements \cite{Hinton2006a,Bottou2007}.  In contrast, deep linear networks gain no expressive power from depth; a shallow network can compute any function that the deep network can, by simply taking $\ws=\wb\wa$.  However, as we see below, the learning dynamics of the deep network is highly \emph{nonlinear}, while the learning dynamics of the shallow network remains linear.  Strikingly, many complex, nonlinear features of learning  appear even in deep linear networks, and do not require neuronal nonlinearities.

As an illustration of the power of deep linear networks to capture learning dynamics even in nonlinear networks, we compare the two learning dynamics in \figprogdiff.  \figprogdiff A shows a low dimensional visualization of the simulated learning dynamics of a multilayered nonlinear neural network trained to predict the properties of a set of items in a semantic domain of animals and plants (details of the neural architecture and training data can be found in \cite{Rogers2004}).  The nonlinear network exhibits a striking, hierarchical progressive differentiation of structure in its internal hidden representations, in which animals versus plants are first distinguished, then birds versus fish, and trees versus flowers, and finally individual items.  This remarkable phenomenon raises important questions about the theoretical principles governing the hierarchical differentiation of structure in neural networks.  In particular, how and why do the network's dynamics and the statistical structure of the input conspire to generate this phenomenon? In \figprogdiff B we mathematically \emph{derive} this phenomenon by finding \emph{analytic} solutions to the nonlinear dynamics of learning in a deep linear network, when that network is exposed to a hierarchically structured semantic domain, 
\begin{figure}
\begin{center}
\includegraphics[width=3.5in]{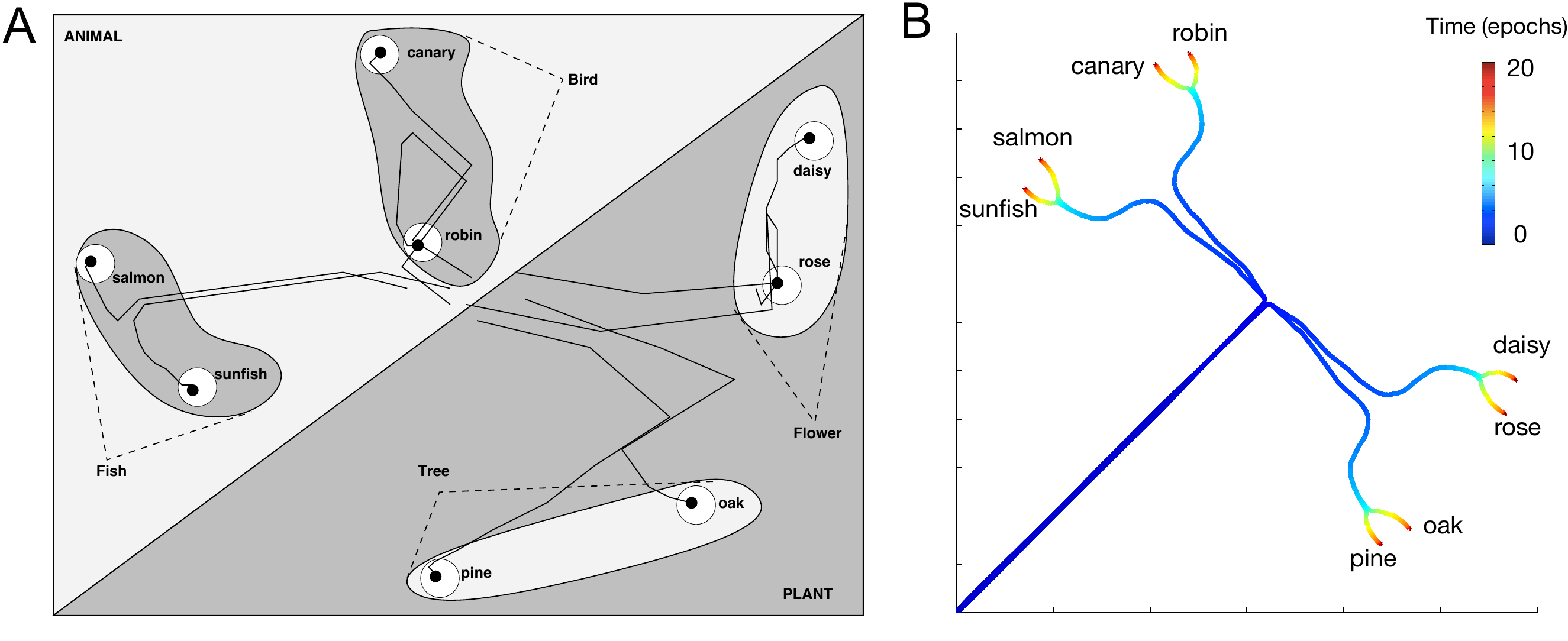}
\end{center}
\caption{(A) A two dimensional multi-dimensional scaling (MDS) visualization of the temporal evolution of internal representations, across developmental time, of a  deep nonlinear neural network studied in \cite{Rogers2004}. Reprinted from Figure 3.9, p. 113 of Ref \cite{Rogers2004}. Copyright \textcopyright ~MIT Press, Permission Pending. (B) An MDS visualization of analytically derived learning trajectories of the internal representations of a deep linear network exposed to a hierarchically structured domain. 
}
\label{fig_prog_diff_numerical_analytic}
\end{figure}
thereby shedding considerable theoretical insight onto the origins of hierarchical differentiation in a deep network.  We present the derivation below, but for now, we note that the resemblance between \figprogdiff A and \figprogdiff B suggests that deep linear networks can form an excellent, analytically tractable model for shedding conceptual insight into the learning dynamics, if not the expressive power, of their nonlinear counterparts.  
%Moreover, as we shall see, deep linear networks generalize in distinct ways from shallow networks. Their hidden layers permit inputs to be converted into an internal representational space, such that similarities between hidden layer activity patterns represent similarities between items. The self-organization of this internal representation yields phenomena that go beyond direct association models, yielding complex generalizations and transient illusory correlations not possible in shallow networks.  Deep linear networks thus constitute a new class of tractable models for studying learning dynamics, patterns of generalization, and conceptual reorganization during development--phenomena that turn out to be distinct from the increasing expressive power conferred by neural nonlinearities and depth.
\section{Acquiring Knowledge}
We now begin an outline of the derivation that leads to \figprogdiff B.   The incremental error corrective process described above can be formalized as online stochastic gradient descent; each time an example $i$ is presented, the weights $\wb$ and $\wa$ are adjusted by a small amount in the direction that most rapidly decreases the squared error $\left\| \y^i - \hat \y^i \right\|^2$, yielding the standard back propagation learning rule
 \beqr
	\Delta \wa  =  \lambda \wb^T \left( \y^i - \hat \y^i  \right) \x^{i T}\label{w_delta_bp},  \, \,
	\Delta \wb  =  \lambda  \left( \y^i - \hat \y^i \right) \h^{i T}, 
\eeqr
where $\lambda$ is a small learning rate.  This incremental update depends {\it only} on experience with a {\it single} item, leaving open the fundamental conceptual question of how and when the accumulation of such incremental updates can extract over developmental time, abstract structures, like hierarchical taxonomies, that are emergent properties of the {\it entire} domain of items and their features. 

We show the extraction of such abstract domain structure is possible provided learning is gradual, with a small learning rate $\lambda$.  In this regime, many examples are seen before the weights appreciably change, so learning is driven by the statistical structure of the domain.  We imagine training is divided into a sequence of learning epochs. In each epoch the above rule is followed for all $P$ examples in random order. Then averaging \eqref{w_delta_bp} over all $P$ examples and taking a continuous time limit gives the mean change in weights per learning epoch,
\beqr
	\tau \ddt \wa & = & \wb^T \left( \sio - \wb \wa \si \right),  \label{wa_avg}\\
	\tau \ddt \wb & = &   \left( \sio - \wb \wa \si \right) \wa^T,\label{wb_avg}
\eeqr
where  
$\si \equiv  E[\x \x^T]$ is an $N_1 \times N_1$ input correlation matrix, 
$\sio \equiv  E[\y \x^T] \label{sio_def}$ is an $N_3 \times N_1$ input-output correlation matrix, and $\tau \equiv \frac{1}{P\lambda}$ (see SI for detailed derivation). Here, $t$ measures time in units of learning epochs; as $t$ varies from 0 to 1, the network has seen $P$ examples corresponding to one learning epoch.   These equations reveal that learning dynamics in even in our simple linear network can be highly complex: the second order statistics of inputs and outputs drives synaptic weight changes through coupled {\it nonlinear} differential equations with up to cubic interactions in the weights. 

\subsection{Explicit solutions from {\normalsize \emph{tabula rasa}}}

These nonlinear dynamics are difficult to solve for arbitrary initial conditions on synaptic weights.  However, we are interested in a particular limit: learning from a state of essentially no knowledge, which we model as small random synaptic weights. To further ease the analysis, we shall assume that the influence of perceptual correlations is minimal ($\si \approx \bf I$). Our fundamental goal, then, is to understand the dynamics of learning in (\ref{wa_avg})-(\ref{wb_avg}) as a function of the input-output correlation matrix $\sio$. The learning dynamics is closely related to terms in the singular value decomposition (SVD) of $\sio$ (\figsvd A),

\beq
	\sio = \ls \sv \rs^T = \sum_{\alpha=1}^{N_1} s_\alpha \uv^\alpha \vv^{\alpha T}, \label{eq_svd}
\eeq
which decomposes any matrix into the product of three matrices. Each of these matrices has a distinct semantic interpretation. 

For example, the $\alpha$'th column $\vv^\alpha$ of the $N_1 \times N_1$ orthogonal matrix $\rs$ can be thought of as an object analyzer vector; it determines the position of item $i$  along an important semantic dimension $\alpha$ in the training set through the component $\vv^\alpha_i$.  To illustrate this interpretation concretely, we consider a simple example dataset with four items (\textit{Canary, Salmon, Oak,} and \textit{Rose}) and five properties (\figsvd). The two animals share the property \textit{can Move}, while the two plants do not. Also each item has a unique property:  \textit{can Fly}, \textit{can Swim},  \textit{has Bark}, and \textit{has Petals}.  For this dataset, while the first row of $\rs^T$ is a uniform mode, the second row, or the second object analyzer vector $\vv^2$, determines where items sit on an \textit{animal-plant} dimension, and hence has positive values for the \textit{Canary} and \textit{Salmon} and negative values for the plants. The other dimensions identified by the SVD are a \textit{bird-fish} dimension, and a \textit{flower-tree} dimension. 

The corresponding $\alpha$'th column $\uv^\alpha$ of the $N_3 \times N_3$ orthogonal matrix $\ls$ can be thought of as a \textit{feature synthesizer} vector for semantic distinction $\alpha$. It's components $\uv^\alpha_m$ indicate the extent to which feature $m$ is present or absent in distinction $\alpha$.  Hence the feature synthesizer $\uv^2$ associated with the \textit{animal-plant} semantic dimension has positive values for \textit{Move} and negative values for \textit{Roots}, as animals typically can move and do not have roots, while plants behave oppositely. Finally the $N_3 \times N_1$ \textit{singular value} matrix $\sv$ has nonzero elements $s_\alpha,\alpha=1,\ldots,N_1$ only on the diagonal, ordered so that $s_1 \geq s_2 \geq \cdots\geq s_{N_1}$.  $s_\alpha$ captures the overall strength of the association between the $\alpha$'th input and output dimensions. The large singular value for the \textit{animal-plant} dimension reflects the fact that this one dimension explains more of the training set than the finer-scale dimensions like \textit{bird-fish} and \textit{flower-tree}. 

\begin{figure}
\begin{center}
\includegraphics[width=3.5in]{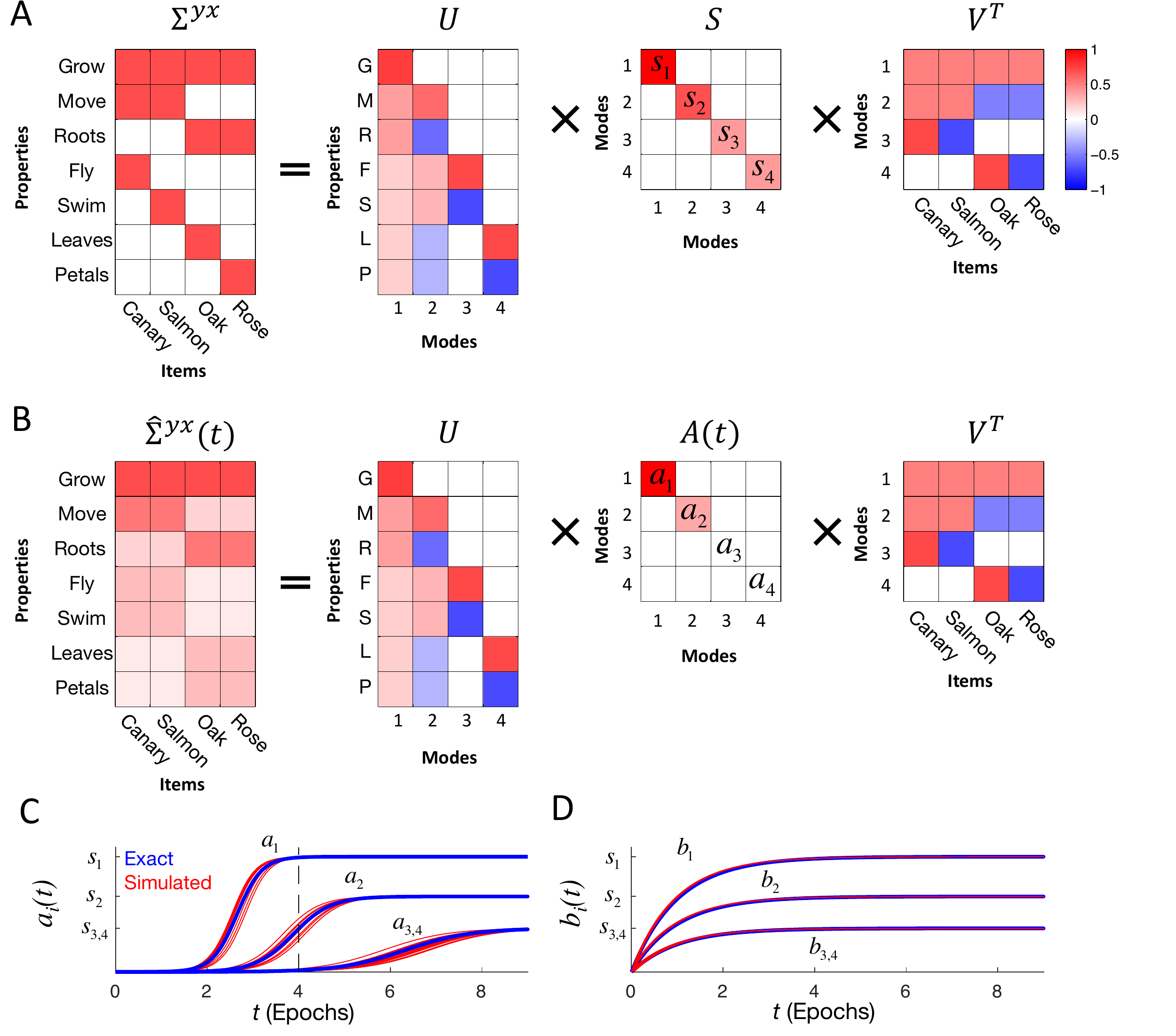}
\end{center}

\caption{(A) Singular value decomposition (SVD) of input-output correlations. Associations between items and their properties are decomposed into modes. Each mode links a set of coherently covarying properties (a column of $\ls$) with a set of coherently covarying items (a row of $\rs^T$). The strength of the mode's covariation is encoded by the singular value of the mode (diagonal element of $\sv$). (B) Network input-output map, analyzed via the SVD. The effective singular  values (diagonal elements of $A(t)$) evolve over time during learning. (C) Time-varying trajectories of the deep network's effective singular values $a_i(t)$. Black dashed line marks the point in time depicted in panel B. (D) Time-varying trajectories of a shallow network's effective singular values $b_i(t)$.
}
\label{fig2}
\end{figure}

Given the SVD of the training set's input-output correlation matrix in \req{eq_svd},  we can now explicitly describe the network's learning dynamics. The network's overall input-output map at time $t$ is a time-dependent version of this SVD decomposition (\figsvd B); it shares the object analyzer and feature synthesizer matrices of the SVD of $\sio$, but replaces the singular value matrix $\sv$ with an effective singular value matrix $\at$,
\beq
	\wb(t) \wa(t) = \ls \at \rs^T =\sum_{\alpha=1}^{N_2} a_\alpha(t) \, \uv^\alpha \vv^{\alpha T}, \label{timecourse}
\eeq
where the trajectory of each effective singular value $a_\alpha(t)$ obeys
\beq
	a_\alpha(t)=\frac{s_\alpha e^{2s_\alpha t/\tau}}{e^{2s_\alpha t/\tau}-1+s_\alpha/a_\alpha^0}. \label{adynam}
\eeq
Eqn.~\ref{adynam} describes a sigmoidal trajectory that begins at some initial value $a_\alpha^0$ at time $t=0$ and rises to $s_\alpha$ as $t\rightarrow \infty$, as plotted in \figsvd C. This solution is applicable when the network begins as a \emph{tabula rasa}, or an undifferentiated state with little initial knowledge, corresponding small random initial weights (see SI for derivation), and it provides an accurate description of the learning dynamics in this regime, as confirmed by simulation in \figsvd C. 

This solution also gives insight into how the internal representations in the hidden layer of the deep network evolve. An exact solution for $\wb$ and $\wa$ is given by 
\beqr
\wa(t)  = {\bf Q} \sqrt{\at}\rs^T, \label{internalrep1} \qquad \wb(t)  =  \ls \sqrt{\at}{\bf Q}^{-1}, \label{internalrep}
\eeqr
where $\bf Q$ is an arbitrary $N_2\times N_2$ invertible matrix (SI Appendix). If initial weights are small, then the matrix $\bf Q$ will be close to orthogonal, i.e., ${\bf Q}\approx \br$ where $\br^T\br=\bf I$. Thus the internal representations are specified up to an arbitrary rotation $\br$. Factoring out the rotation, the hidden representation of item $i$ is 
\beq
h^\alpha_i=\sqrt{a^\alpha(t)}\vv_i^\alpha.
\label{eq:hiddenrep}
\eeq
Thus internal representations develop over time by projecting inputs onto more and more input-output modes as they are learned. 

The shallow network has a solution of analogous form,
$ \ws(t) =\sum_{\alpha=1}^{\min(N_1,N_3)} b_\alpha(t) \, \uv^\alpha \vv^{\alpha T}, $
but now each singular value evolves as
\beq
	b_\alpha(t) = s_\alpha \left(1-e^{-t/ \tau}\right) + b_\alpha^0e^{-t/\tau}. \label{bdynam}
\eeq
In contrast to the deep network's sigmoidal trajectory, Eqn.~\ref{bdynam} describes a simple exponential approach from the initial value $b^0_\alpha$ to $s_\alpha$, as plotted in \figsvd D. Hence depth fundamentally changes the dynamics of learning, yielding several important consequences below. 

\subsection{Rapid stage like transitions due to depth} ~
We first compare the time-course of learning in deep versus shallow networks as revealed in Eqns.~\req{adynam} and~\req{bdynam}.  For the deep network, beginning from a small initial condition $a_\alpha^0=\epsilon,$ each mode's effective singular value $a_\alpha(t)$ rises to within $\epsilon$ of its final value $s_\alpha$  in time
\beq
	t(s_\alpha,\epsilon) = \frac{\tau}{s_\alpha}\ln\frac{s_\alpha}{\epsilon}\label{tauovs_deep}
\eeq
 in the limit $\epsilon\rightarrow 0$ (SI Appendix). This is $O(1/s_\alpha)$ up to a logarithmic factor. Hence modes with stronger explanatory power, as quantified by the singular value, are learned more quickly. Moreover, when starting from small initial weights, the sigmoidal transition from no knowledge of the mode to perfect knowledge can be arbitrarily sharp.  Indeed the ratio of time spent in the sigmoidal transition regime to the ratio of time spent before making the transition can go to infinity as the initial weights go to zero (see SI Appendix).  Thus rapid stage like transitions in learning can be prevalent even in deep linear networks. 
 
By contrast, the timescale of learning for the shallow network is
\beq
	t(s_\alpha,\epsilon) = \tau\ln\frac{s_\alpha}{\epsilon}, \label{tauovs_shallow}
\eeq
which is $O(1)$ up to a logarithmic factor. Hence in a shallow network, the timescale of learning a mode depends only weakly on its associated singular value.  Essentially all modes are learned at the same time, with an exponential rather than sigmoidal learning curve. 

\subsection{Progressive differentiation of hierarchical structure}

We are now almost in a position to explain how we analytically derived the result in \figprogdiff B.  The only remaining ingredient is a mathematical description of the training data. Indeed the numerical results in \figprogdiff A arose from a toy-training set, making it difficult to understand which aspects the data were essential for the hierarchical learning dynamics.  Here, we introduce a probabilistic generative model for hierarchically structured data, in order to move beyond toy datasets to extract general principles of how statistical structure impacts learning. 

We use a generative model (introduced in \cite{Kemp2004}) that mimics the process of evolution to create a dataset with explicit hierarchical structure (see SI Appendix).  In the model, each feature diffuses down an evolutionary tree (\fighiersvd A), with a small probability of mutating along each branch.  The items lie at the leaves of the tree, and the generative process creates a hierarchical similarity matrix between items such that items with a more recent common ancestor on the tree are more similar to each other (\fighiersvd B).  We analytically computed the SVD of this hierarchical dataset and we found that the object analyzer vectors, which can be viewed as functions on the leaves of the tree in \fighiersvd C respect the hierarchical branches of the tree, with the larger (smaller) singular values corresponding to broader (finer) distinctions. Moreover, in \fighiersvd A we have artificially labelled the leaves and branches of the evolutionary tree with organisms and categories that might reflect a natural realization of this evolutionary process.   

\begin{figure}
\begin{center}
\includegraphics[width=3.5in]{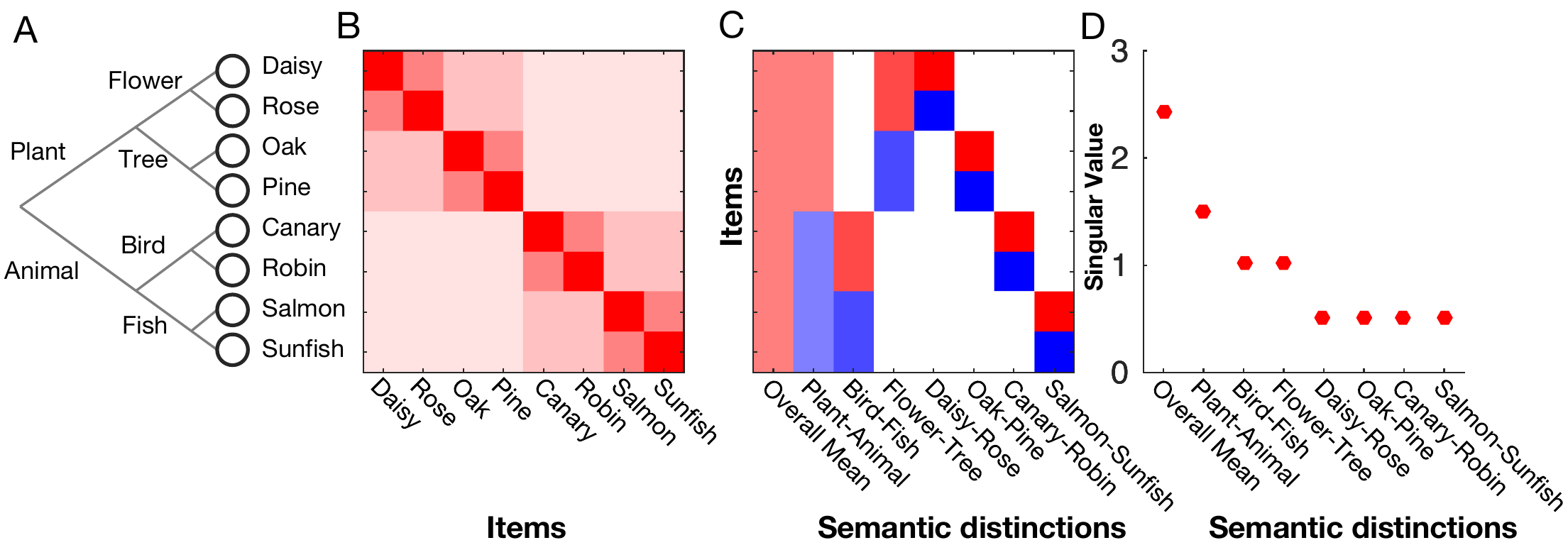}
\end{center}
\caption{Hierarchy and the SVD. (A) A domain of eight items with an underlying hierarchical structure. (B) The correlation matrix of the features of the items. (C) Singular value decomposition of the correlations reveals semantic distinctions that mirror the hierarchical taxonomy. This is a general property of the SVD of hierarchical data. (D) The singular values of each semantic distinction reveal its strength in the dataset, and control when it is learned.}
\label{hierarchical_svd_fig}
\end{figure}

Now, inserting the singular values in \fighiersvd D (and SI Appendix) into the deep learning dynamics in Eq. \ref{adynam} to obtain the time-dependent singular values $a^\alpha(t)$, and then inserting these along with the object analyzers vectors $\vv^\alpha$ in \fighiersvd C into Eq. \ref{eq:hiddenrep}, we obtain a complete analytic derivation of the evolution of internal representations over developmental time in the deep network.  An MDS visualization of these evolving hidden representation then yields \figprogdiff B, which qualitatively recapitulates the much more complex network and dataset that led to \figprogdiff A.   In essence, this analysis completes a mathematical proof that the striking progressive differentiation of hierarchical observed in \figprogdiff \, is an inevitable consequence of deep learning dynamics, even in linear networks, when exposed to hierarchically structured data.  The essential intuition is that dimensions of feature variation across items corresponding to broader (finer) hierarchical distinctions have stronger (weaker) statistical structure, as quantified by the singular values of the training data, and hence these dimensions are learned faster (slower), leading to waves of differentiation in a deep, but not a shallow network.  Such a pattern of hierarchical differentiation has long been argued to apply to the conceptual development of infants and children \cite{Keil1979, Mandler1993,Inhelder1958,Siegler1976}.% (but see \sg{[cite**]}).

\subsection{Illusory Correlations}

Another intriguing aspect of semantic development is that children sometimes attest to false beliefs (i.e. worms have bones \cite{Carey1985}) that could not have been learned through direct experience. These errors challenge simple associationist accounts of semantic development that would predict a steady, monotonic accumulation of information about individual properties \cite{Murphy1985,Carey1985,Gelman1990,Keil1991}. Yet as shown in \figilluscorrs, the network's knowledge of individual properties exhibits complex, non-monotonic trajectories over the course of learning. The overall prediction for a property is a sum of contributions from each mode, where the specific contribution of mode $\alpha$ to an individual feature $m$ for item $i$ is $a_\alpha(t)\uv^\alpha_m \vv^\alpha_i$. In the example of \figilluscorrs A, the first two modes make a positive contribution while the third makes a negative one, yielding the inverted U-shaped trajectory. 

Indeed, any property-item combination for which $\uv^\alpha_m \vv^\alpha_i$ takes different signs across different $\alpha$ will exhibit a non-monotonic learning curve, making such curves a frequent occurrence even in a fixed, unchanging environment. 
%The impact of depth is to prolong and amplify these illusory correlations (\figilluscorrs B).  
In a deep network, two modes with singular values that differ by $\Delta$ will have an interval in which the first is learned but the second is not. The length of this developmental interval is roughly $O(\Delta)$ (SI Appendix). Moreover, the rapidity of the deep network's stage-like transitions further accentuates the non-monotonic learning of individual properties.  This behavior, which may seem hard to reconcile with an incremental, error-corrective learning process, is a natural consequence of minimizing  global rather than local prediction error: the fastest possible improvement in predicting all properties across all items sometimes results in a transient increase in the size of errors on specific items and properties. Every property in a shallow network, by contrast, monotonically approaches its correct value and therefore shallow networks provably never exhibit illusory correlations (SI Appendix).  %Thus transient illusory correlations need not reflect defective learning, but rather may actually be an indirect consequence of global optimality in semantic learning, as it is in this simple network.  

\begin{figure}
\begin{center}
\includegraphics[width=3.2in]{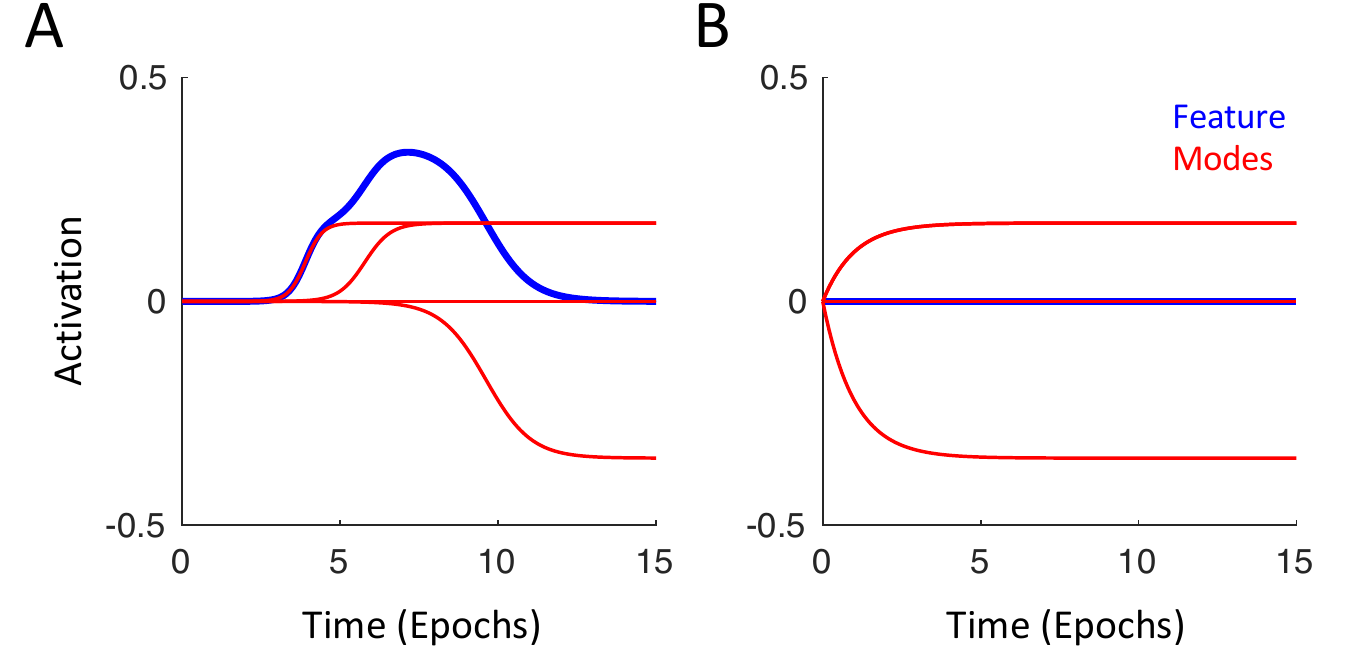}
\end{center}
\caption{Illusory correlations during learning. (A) Predicted value (blue) of feature ``Can Fly'' for item ``Salmon'' over the course of learning in a deep network (dataset as in \figsvd).  The contributions from each input-output mode are shown in red. (B) The predicted value and modes for the same feature in a shallow network. 
}
\label{fig3}
\end{figure}

\section{Organizing and Encoding Knowledge}
We now turn from the dynamics of learning to its final outcome.  When exposed to a variety of items and features interlinked by an underlying hierarchy, for instance, what categories naturally emerge? Which items are particularly representative of a categorical distinction? And how is the structure of the domain internally represented?

\subsection{Category membership, typicality, and prototypes}

A long observed empirical finding is that category membership is not simply a logical, binary variable, but rather a {\it graded} quantity, with some objects being more or less typical members of a category (i.e. a \textit{sparrow} is a more typical bird than a \textit{penguin}).  Indeed, such graded judgements of category membership are both highly reproducible across individuals \cite{Rosch1975,Barsalou1985} and moreover correlate with performance on a range of tasks:  subjects more quickly verify the category membership of more typical items \cite{Rips1973,Murphy1985a}, more frequently recall typical examples of a category  \cite{Mervis1976}, and more readily extend new information about typical items to all members of a category \cite{Rips1975,Osherson1990}.  Our theoretical framework provides a natural mathematical definition of item typicality that both explains how it emerges from the statistical structure of the environment and improves task performance. 

Indeed, a natural notion of the typicality of an item $i$ for a categorical distinction $\alpha$ is simply the quantity $\vv^\alpha_i$ in the corresponding object analyzer vector.  To see why this is natural, note that after learning, the neural network's internal representation space has a semantic distinction axis $\alpha$, and each object $i$ is placed along this axis at a coordinate proportional to $\vv^\alpha_i$, as seen in Eq. \req{eq:hiddenrep}.  Thus according to our definition, extremal points along this axis are the most typical members of a category.  For example, if $\alpha$ corresponds to the bird-fish axis, objects $i$ with large positive $\vv^\alpha_i$ are typical birds, while objects $i$ with large negative $\vv^\alpha_i$ are typical fish.  Also, the contribution of the network's output to feature neuron $m$ in response to object $i$, from the hidden representation axis $\alpha$ alone is given by 
\beq
	\hat \y^\alpha_m \leftarrow \uv^\alpha_m s_\alpha \vv^\alpha_i. \label{typicality_feat}
\eeq
Hence under our definition of typicality, an item $i$ that is more typical than another other item $j$ will have $|\vv^\alpha_i|>|\vv^\alpha_j|$, and thus will necessarily have a larger response magnitude under Eq. \req{typicality_feat}. Any performance measure which is monotonic in the response will therefore increase for more typical items under this definition.  Thus our definition of item typicality is both a mathematically well defined function of the statistical structure of experience, through the SVD, and proveably correlates with task performance in our network.

Several previous attempts at defining the typicality of an item involve computing a weighted sum of category specific features present or absent in the item \cite{Mervis1982,Davis2014,Rosch1975,Rosch1976,Rosch1978}. For instance, a \textit{sparrow} is a more typical bird than a \textit{penguin} because it shares more relevant features (\textit{can fly}) with other birds. However, the specific choice of which features are relevant--the weights in the weighted sum of features--has often been heuristically chosen and relied on prior knowledge of which items belong to each category \cite{Rosch1975,Davis2014}. Our definition of typicality can also be described in terms of a weighted sum of an object's features, but the weightings are {\it uniquely} fixed by the statistics of the entire environment through the SVD (see SI Appendix): 
\beqr
	\vv^\alpha_i & = & \frac{1}{Ps_\alpha} \sum_{m=1}^{N_3}  \uv^\alpha_m \ov^{i}_m,   \label{v_typ} %\\
	%\uv^\alpha_m & = &  \frac{1}{s_\alpha} \sum_{i=1}^{N_1}  \vv^\alpha_i \fv^{m}_i \label{u_typ}
\eeqr
which holds for all $i,m,$ and $\alpha$. Here, item $i$ is defined by its feature vector $\ov^i\in R^{N_3}$, where component $\ov^i_m$ encodes the value of its $m^{th}$ feature. Thus the typicality $\vv^\alpha_i$ of item $i$ in distinction $\alpha$ can be computed by taking a weighted sum of the components of its feature vector $\ov^i$, where the weightings are precisely the coefficients of the corresponding feature synthesizer vector $\uv^\alpha$ (scaled by the reciprocal of the singular value). The neural geometry of Eq. \ref{v_typ} is illustrated in \figtypgeom \,when $\alpha$ corresponds to the bird-fish categorical distinction. 

In many theories of typicality, the particular weighting of object features corresponds to a prototypical object \cite{Rosch1978,Murphy2002}, or the best example of a particular category. Such object prototypes are often obtained by a weighted average over the feature vectors for the objects in a category (i.e. averaging together the features of all birds, for instance, will give a set of features they share).  However, such an average relies on prior knowledge of the extent to which an object belongs to a category.  Our theoretical framework also yields a natural notion of object prototypes as a weighted average of object feature vectors, but unlike many other frameworks, it yields a {\it unique} prescription for the object weightings in terms of the statistical structure of the environment through the SVD (SI Appendix):
\beqr
	%\vv^\alpha_i & = & \frac{1}{s_\alpha} \sum_{m=1}^{N_3} \fv^{m}_i \uv^\alpha_m, \label{v_prot} \\
	\uv^\alpha_m & = &  \frac{1}{Ps_\alpha}\sum_{i=1}^{N_1}  \vv^\alpha_i \ov^{i}_m. \label{u_prot}
\eeqr
Thus the feature synthesizer $\uv^\alpha$, can itself be thought of as a category prototype for distinction $\alpha$, as it can be obtained through a weighted average of {\it all} the object feature vectors $\ov^{i}$, where the weighting of object $i$ is none other than the {\it typicality}  $\vv^\alpha_i$ of object $i$ in distinction $\alpha$.  In essence, each element $\uv^\alpha_m$ of the prototype vector signifies how important feature $m$ is for distinction $\alpha$ (\figtypgeom). 

In summary, a beautiful and simple duality between item typicality and category prototypes arises as an emergent property of the learned internal representations of the neural network.  The typicality of an item is determined by the projection of that item's feature vector onto the category prototype in \req{v_typ}. And the category prototype is an average over all object feature vectors, weighted by their typicality in \req{u_prot}.  Moreover, in any categorical distinction $\alpha$, the most typical items $i$ and the most important features $m$ are determined by the {\it extremal} values of $\vv^\alpha_i$ and $\uv^\alpha_m$.

\begin{figure}
\begin{center}
\includegraphics[width=3.4in]{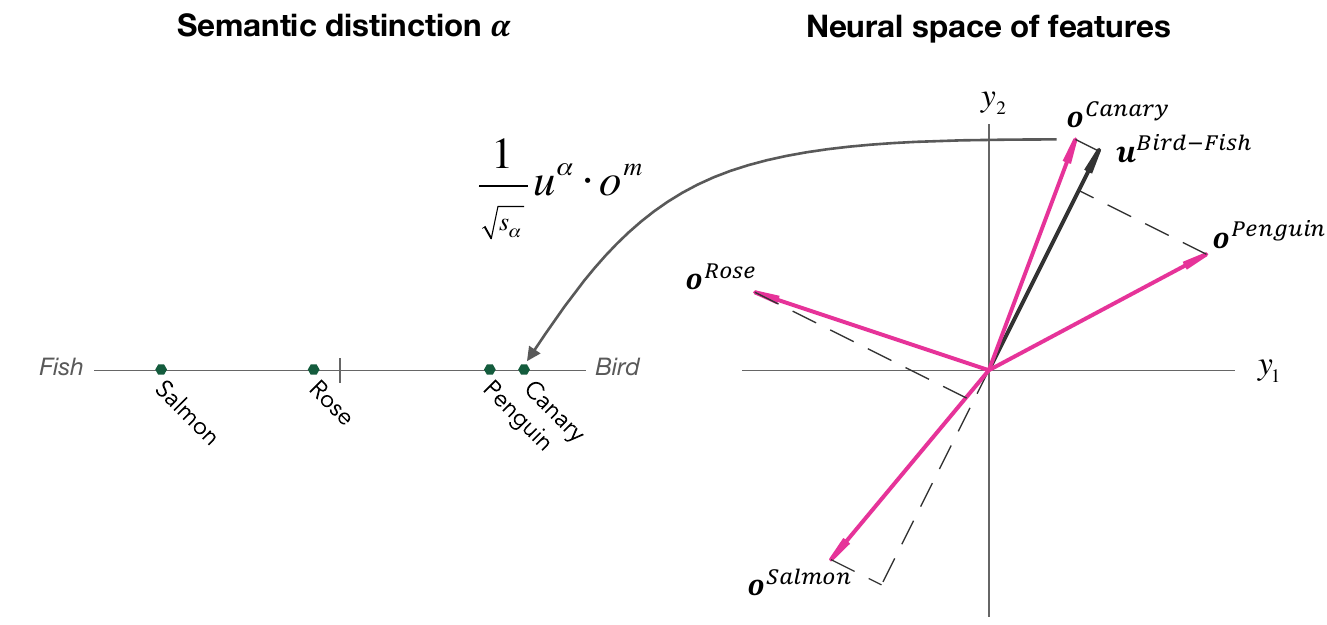}
\end{center}

\caption{The geometry of item typicality.  For a semantic distinction $\alpha$ (in this case $\alpha$ is the bird-fish distinction) the object analyzer vector $\vv^\alpha_i$ arranges objects $i$ along an internal neural representation space where the most typical birds take the extremal positive coordinates, and the most typical fish take the extremal negative coordinates. Objects like a rose, that is neither a bird nor a fish, are located near the origin on this axis.  Positions along the neural semantic axis can also be obtained by computing the inner product between the feature vector $\ov^i$ for object $i$ and the feature synthesizer  $\uv^\alpha$ as in \req{v_typ}. Moreover $\uv^\alpha$ can be thought of as a category prototype for semantic distinction $\alpha$ through \req{u_prot}.
}
\label{fig5_simplified}
\end{figure}

\subsection{Category coherence}

The categories we naturally learn are not arbitrary, but instead are in some sense coherent, and efficiently represent the structure of the world \cite{Rosch1975,Rosch1978,Murphy1985}.  For example, the set of things that are \textit{are Red} and \textit{cannot Swim}, is a well defined category, but intuitively is not as coherent as the category of \textit{Dogs}; we naturally learn, and even name, the latter category, but not the former. When is a category learned at all, and what determines its coherence? An influential proposal \cite{Rosch1975,Rosch1978} suggested that coherent categories consist of tight clusters of items that share many features, and moreover are highly distinct from other categories with different sets of shared features.   Such a definition, as noted in \cite{Murphy2002, Murphy1985, Keil1991} can be circular: to know which items are category members, one must know which features are important for that category, and conversely, to know which features are important, one must know which items are members. Thus a mathematical definition of category coherence, as a function of the statistical structure of the environment, that is proveably related to the learnability of categories by neural networks, has remained elusive.

Here we provide such a definition for a simple model of disjoint categories, and demonstrate how neural networks can cut through the Gordian knot of circularity. Our definition and theory is motivated by, and consistent with, prior network simulations exploring notions of category coherence through the coherent covariation of features \cite{Rogers2004}.        
\begin{figure}
\begin{center}
\includegraphics[width=3.5in]{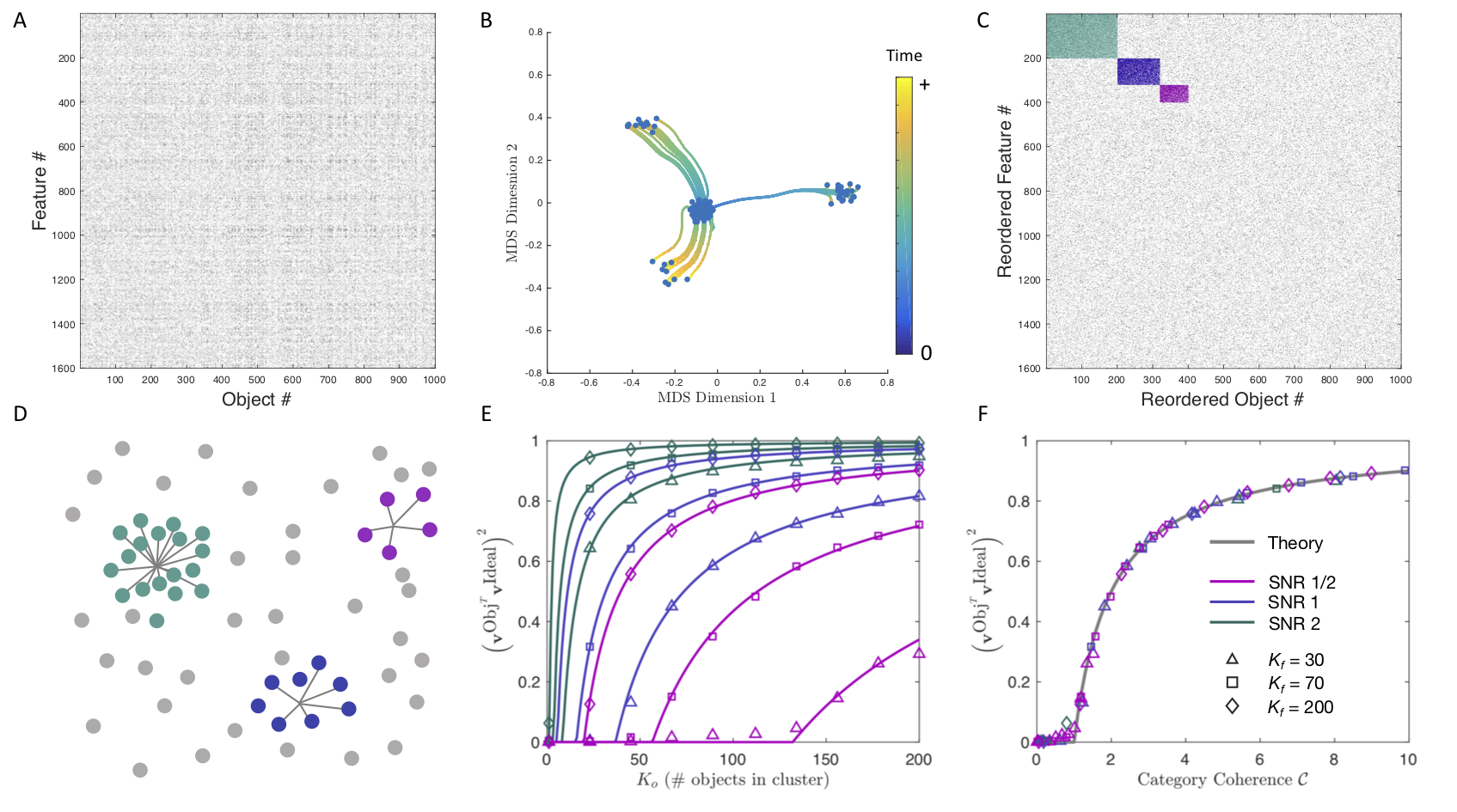}
\end{center}
\caption{The discovery of disjoint categories buried in noise. (A) A data set of $N_0 = 1000$ items and $N_f = 1600$ features, with no discernible visible structure.  (B) Yet when a deep linear network learns to predict the features of items, an MDS visualization of the evolution of its internal representations reveals $3$ clusters. (C)  By computing the SVD of the product of synaptic weights $\wb \wa$, we can extract the network's object analyzers $\vv^\alpha$ and feature synthesizers $\uv^\alpha$, finding $3$ with large singular values $s_\alpha$, for $\alpha=1,\dots,3$.  Each of these $3$ object analyzers $\vv^\alpha$ and feature synthesizers $\uv^\alpha$ takes large values on a subset of items and features respectively, and we can use them to reorder the rows and columns of (A) to obtain  (C).  This re-ordering reveals underlying structure in the data corresponding to $3$ disjoint categories, such that if a feature and item belong to a category, the feature is present with a high probability $p$, whereas if it does not, it appears with a low probability $q$.  (D) Thus intuitively, the dataset corresponds to $3$ clusters buried in a noise of irrelevant objects and features. (E) Performance in recovering one such category can be measured by computing the correlation coefficients between the object analyzer and feature synthesizer returned by the network to the ideal object analyzer $\vv^\text{Ideal}$ and ideal feature synthesizer $\uv^\text{Ideal}$ that take nonzero values on the items and features, respectively, that are part of the category, and are zero on the rest of the items and features. This learning performance, for the object analyzer, is shown for various parameter values. Solid curves are analytical theory derived from a random matrix analysis (SI Appendix)  and data points are obtained from simulations. (F)  All performance curves in (E) collapse onto a {\it single} theoretically predicted, universal learning curve, when measured in terms of the category coherence defined in Eq. \ref{eq:catcohdisj}.}
\label{fig_categ_coh}
\end{figure}

Consider for example, a dataset consisting of $N_o$ objects and $N_f$ features.  Now consider a category consisting of a subset of $K_f$ features that tend to occur with high probability $p$ in a subset of $K_o$ items, whereas a background feature occurs with a lower probability $q$ in a background item $p$ when either are not part of the category. For what values of $K_f$, $K_0$, $p$, $q$, $N_f$ and $N_0$ can such a category be learned, and if so, how accurately?  \figcategcoh A-D illustrates, for example, how a neural network can extract $3$ such categories buried in the background noise.  We see in \figcategcoh E that the performance of category extraction increases as the number of items $K_0$ and features $K_f$ in the category increases, and also as the signal-to-noise ratio, or $\text{SNR} \equiv \frac{(p-q)^2}{q(1-q)}$ increases.  Through random matrix theory (SI Appendix), we show that performance depends on the various parameters $\it{only}$ through a category coherence variable
\begin{equation}
\mathcal C = \text{SNR} \frac{K_o K_f}{\sqrt{N_o N_f}}.
\label{eq:catcohdisj}
\end{equation}
When the performance curves in \figcategcoh E are re-plotted with category coherence $\mathcal C$ on the horizontal axis, all the curves collapse onto a single \emph{universal} performance curve shown in \figcategcoh F.  We derive a mathematical expression for this curve in SI Appendix.  It displays an interesting threshold behavior: if the coherence $\mathcal{C} \leq 1$, the category is not learned at all, and the higher the coherence above this threshold, the better the category is learned.  

This threshold is strikingly permissive.  For example, at $\text{SNR}=1$,  it occurs at $K_0 K_f = \sqrt{N_0 N_f}$.  Thus in a large environment of $N_o=1000$ objects and $N_f=1600$ features, as in \figcategcoh A, a small category of $40$ objects and $40$ features can be easily learned, even by a simple deep linear network.  Moreover, this analysis demonstrates how the deep network solves the problem of circularity described above by simultaneously bootstrapping the learning of object analyzers and feature synthesizers in its synaptic weights.  Finally, we note that the definition of category coherence in Eq. \req{eq:catcohdisj} is qualitatively consistent with previous notions; coherent categories consist of large subsets of items possessing, with high probability, large subsets of features that tend not to co-occur in other categories.  However, our quantitative definition has the advantage that it proveably governs category learning performance in a neural network.

\vspace{-.17in}
\subsection{Basic categories}

Closely related to category coherence, a variety of studies of naming have revealed a privileged role for \textit{basic} categories at an intermediate level of specificity (i.e. \textit{Bird}), compared to superordinate (i.e. \textit{Animal}) or subordinate (i.e. \text{Robin}) levels. At this basic level, people are quicker at learning names \cite{Anglin1977,Mervis1976a}, prefer to generate names \cite{Mervis1976a}, and are quicker to verify the presence of named items in images \cite{Mervis1976a,Murphy1985a}. % \as{Jolicoeur et al., 1984; lin et al. 1997; murphy and brownell 1985; murphy \& smith 1982; smith et al 1978; rosch et al 1976; tanaka and taylor 1991}, 
We note that basic level advantages typically involve naming tasks done at an older age, and so need not be inconsistent with progressive differentiation of categorical structure from superordinate to subordinate levels as revealed in preverbal cognition \cite{Keil1979,Mandler1991,Mandler1993,Inhelder1958,Siegler1976,Rogers2004}. Moreover, some items are named more frequently than others, and these frequency effects could contribute to a basic level advantage \cite{Rogers2004}. However in artificial category learning experiments where frequencies are tightly controlled, basic level categories are still often learned first \cite{Corter1992}. What statistical properties of the environment could lead to this basic level effect? While several properties have been put forth in the literature \cite{Rosch1976,Corter1992,Murphy1985a,Mervis1982}, a mathematical function of environmental structure that proveably confers a basic level advantage to neural networks has remained elusive. 

Here we provide such a function by generalizing the notion of category coherence $\mathcal C$ in the previous section to hierarchically structured categories.  Indeed, in any dataset containing strong categorical structure, so that its singular vectors are in one to one correspondence with categorical distinctions, we simply propose to \textit{define} the coherence of a category by the associated singular value.  This definition has the advantage of obeying the theorem that more coherent categories are learned faster, through Eq. \req{adynam}.  Moreover, we show in SI Appendix that this definition is consistent with that of category coherence $\mathcal C$ defined in Eq. \req{eq:catcohdisj} for the special case of disjoint categories. However, for hierarchically structured categories as in \fighiersvd, this singular value definition always predicts an advantage for superordinate categories, relative to basic or subordinate. 

Is there an alternate statistical structure for hierarchical categories that confers high category coherence at lower levels in the hierarchy?  We exhibit two such structures in \figbasiclvl.   More generally, in the SI Appendix, we analytically compute the singular values at each level of the hierarchy in terms of the similarity structure of items.   We find these singular values are a weighted sum of within cluster similarity minus between cluster similarity for all levels below, weighted by the fraction of items that are descendants of that level.   If at any level, between cluster similarity is negative, that detracts from the coherence of superordinate categories, contributes strongly to the coherence of categories at that level, and does not contribute to subordinate categories. 
\begin{figure}
\begin{center}
\includegraphics[width=3.5in]{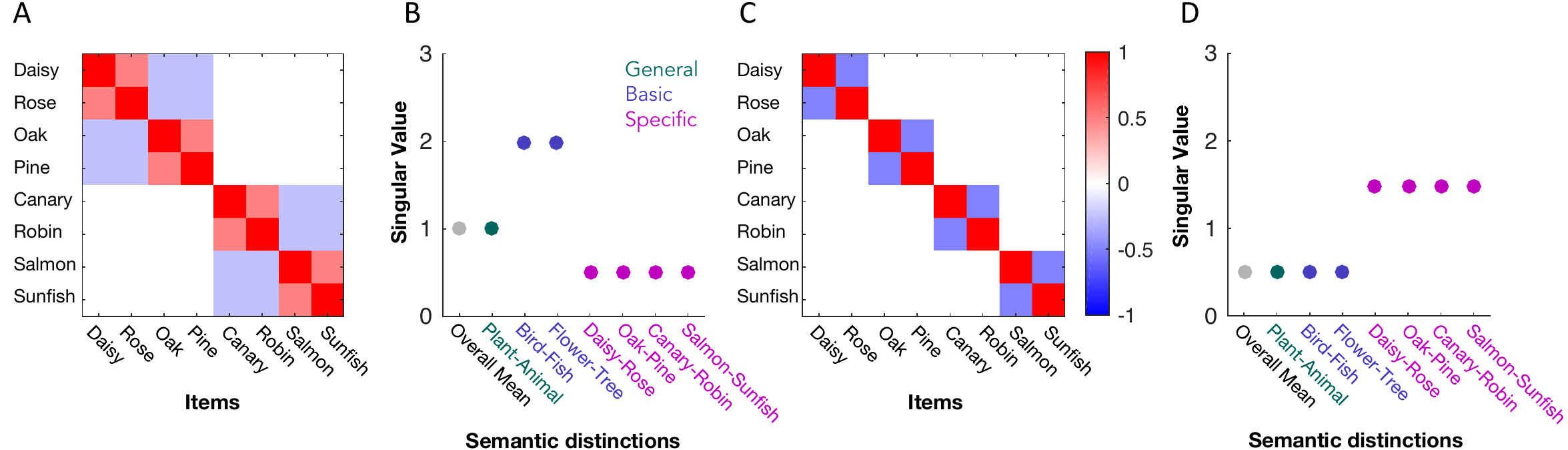} % Can switch to basic_level_figure_horiz.pdf for horizontal version
\end{center}
\caption{From hierarchical similarity structure to category coherence.  (A) A hierarchical similarity structure over objects in which categories at the basic level are very different from each other due to a negative similarity. (B) For such similarity structure,  basic level categorical distinctions acquire larger singular values, or category coherence, and therefore gain an advantage in both learning and in task performance. (C) Now subordinate categories are very different from each other through negative similarity. (D) Consequently, subordinate categories gain a coherence advantage.  See SI Appendix for analytic formulas relating  similarity structure to category coherence.}
\label{fig_basic_level}
\end{figure}

Thus the singular value based definition of category coherence is qualitatively consistent with prior intuitive notions. For instance, paraphrasing Keil (1991), coherent categories are clusters of tight bundles of features separated by relatively empty spaces \cite{Keil1991}.  Also,  consistent with \cite{Murphy1985, Keil1991, Murphy2002}, we note that we cannot judge the coherence of a category without knowing about its relations to all other categories, as singular values are a complex emergent property of the entire environment.  But going beyond past intuitive notions, our quantitative definition of category coherence based on singular values enables us to prove that coherent categories are most easily and quickly learned, and also proveably provide the most accurate and efficient linear representation of the environment, due to the global optimality properties of the SVD (see SI Appendix for details).

\subsection{Discovering and representing explicit structures}
\label{hierarchical_data}

While we have focused on hierarchical structure, the world may contain many different types of abstract structures. How are these different structures learned and encoded by neural networks?  A convenient formalization of environmental structure can be specified in terms of a probabilistic graphical model (PGM), defined by a graph over items (\figstructform \, top) that can express a variety of  structural forms underlying a domain, including clusters, trees, rings, grids, orderings, and hierarchies.  Features are assigned to items by independently sampling from the PGM (see \cite{Kemp2008} and SI Appendix), such that nearby items in the graph are more likely to share features.  For each of 
\begin{figure}
\centering
\includegraphics[width=0.48\textwidth]{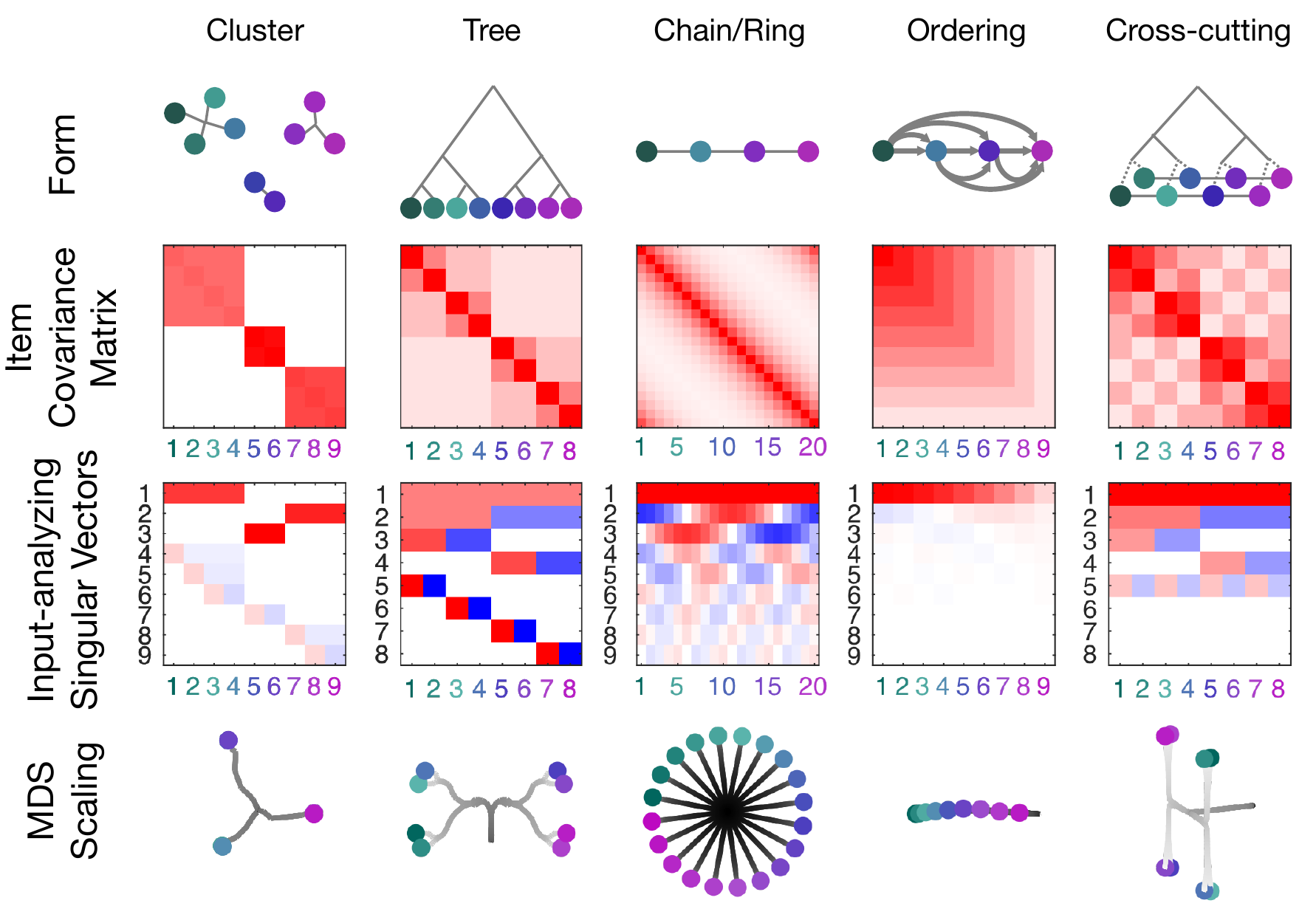} % Or fig4_resized_mds_closedcirc_crosscut_v2 for a version with a hierarchical example
\caption{Representation of explicit structural forms in a neural network. Each column shows a different structure.  The first four columns correspond to pure structural forms, while the final column has cross-cutting structure. First row: The structure of the data generating probabilistic graphical model (PGM). Second row: The resulting item covariance matrix arising from either data drawn from the PGM (first four columns) or designed by hand (final column).  Third row: The input-analyzing singular vectors that will be learned by the linear neural network. Each vector is scaled by its singular value, showing its importance to representing the covariance matrix. Fourth row: MDS view of the development of internal representations.}
\label{fig4}
\end{figure}
these structural forms, in the limit of a large number of features, we computed the item-item covariance matrices (\figstructform \, second row), object analyzer vectors (\figstructform \, third row) and singular values of the resultant input-output correlation matrix, and we employed them in our learning dynamics in Eq. \ref{adynam} to compute the development of the network's internal representations through Eq. \ref{eq:hiddenrep}.  These evolving hidden representations are shown in (\figstructform \, bottom).  Overall, this approach yields several insights into how distinct structural forms, through their different statistics, drive learning in a deep network, as summarized below:

\begin{description}
\item[Clusters.] Graphs that break items into distinct clusters give rise to block-diagonal constant matrices, yielding object-analyzer vectors that pick out cluster membership.
\item[Trees.] Tree graphs give rise to ultrametric covariance matrices, yielding object-analyzer vectors that are tree-structured wavelets that mirror the underlying hierarchy \cite{Khrennikov2005,Murtagh2007a}. %For example, balanced binary trees yield object-analyzer vectors that are Haar wavelets, each one corresponding to a binary distinction in the graph \cite{Murtagh2007a}.
\item[Rings and Grids.] Ring-structured graphs give rise to circulant covariance matrices, yielding object-analyzer vectors that are Fourier modes ordered from lowest to highest frequency \cite{Gray2005}. 
\item[Orderings.] Graphs that transitively order items yield highly structured, but non-standard, covariance matrices whose object analyzers encode the ordering.
%\item[Hierarchies.] The object-analyzer vectors separate hierarchies in a manner similar to trees, but with additional items at each branching point in the hierarchy.
\item[Cross-cutting Structure.] Real-world domains need not have a single underlying structural form. For instance, while some features of animals and plants generally follow a hierarchical structure, other features like \textit{male} and \textit{female} can link together hierarchically disparate items. Such cross-cutting structure can be orthogonal to the hierarchical structure, yielding object-analyzer vectors that span hierarchical distinctions.
\end{description}

In essence, these results reflect an analytic link between two very popular, but different, methods of capturing structure: PGM's and deep networks.  This general analysis transcends the particulars of any one dataset, and shows how different abstract structures become embedded in the internal representations of a deep neural network.

\section{Deploying Knowledge: Inductive Projection}

Over the course of development, the knowledge acquired by children powerfully reshapes their inductions upon encountering novel items and properties \cite{Carey1985,Murphy2002}. For instance, upon learning a novel fact (e.g., ``a canary is warm-blooded'') children extend this new knowledge to related items, as revealed by their answers to questions like "is a robin warm-blooded?" Studies of inductive projection have shown that children's answers to such questions change over the course of development \cite{Carey1985,Carey2011,Keil1991,Murphy2002,Gelman1990a}, generally becoming more specific with age.  For example, young children may readily project the novel property of warm-blooded to distantly related items, while older children will only project it to more closely related items.  How could such changing patterns of inductive generalization arise in a neural network? Here, building upon previous network simulations of inductive projection \cite{Rumelhart1993,Rogers2004}, we show analytically that deep networks exposed to hierarchically structured data, naturally yield progressively narrowing patterns of inductive projection across development.

\begin{figure}
\begin{center}
\includegraphics[width=3.4in]{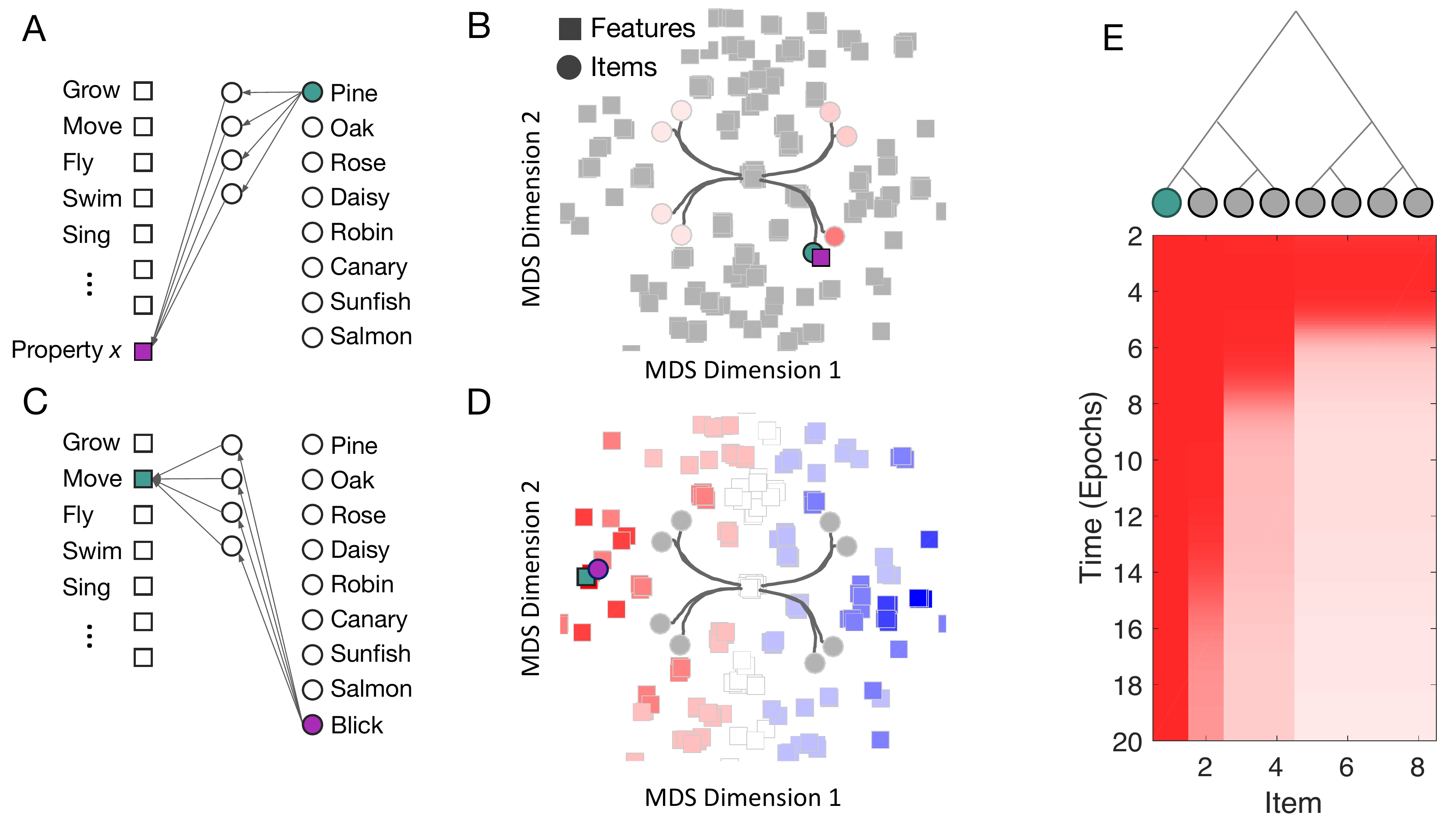}
\end{center}
\caption{The neural geometry of inductive generalization. (A) A novel feature (property \textit{x}) is observed for a familiar item (i.e. "a pine has property \textit{x}"). (B) Learning assigns the novel feature a neural representation in the hidden layer of the network that places it in semantic similarity space near the object which possesses the novel feature. The network then inductively projects that novel feature to other familiar items (e.g. "Does a rose have property \textit{x}?") only if their hidden representation is close in neural space. (C) A novel item (a \textit{blick}) possesses a familiar feature (i.e. "a blick can move"). (D) Learning assigns the novel item a neural representation in the hidden layer that places it in semantic similarity space near the feature possessed by the novel item. Other features are inductively projected to that item (e.g., "Does a blick have wings?") only if their hidden representation is close in neural space. (E) Inductive projection of a novel property ("a pine has property \textit{x}") over learning. As learning progresses, the neural representations of items become progressively differentiated, yielding progressively restricted projection of the novel feature to other items. Here the pine can be thought of as the left-most item node in the tree.}
\label{geom_of_gen}
\end{figure}

Consider the act of learning that a familiar item has a novel feature (e.g. ``a pine has property \textit{x}''). To accommodate this knowledge, new synaptic weights must be chosen between the familiar item \textit{pine} and the novel property \textit{x} (\figinductgen A), without disrupting prior knowledge of items and their properties already stored in the network. This may be accomplished by adjusting {\it only} the weights from the hidden layer to the novel feature so as to activate it appropriately. %while leaving the weights from the known item to the hidden layer unchanged (as these weights play a role in many other feature associations for this item). 
With these new weights established, inductive projections of the novel feature to other familiar items (e.g. ``does a rose have property \textit{x}?'') naturally arise by querying the network with other inputs. If a novel property $m$ is ascribed to a familiar item $i$, the inductive projection of this property to any other item $j$ is given by (see SI Appendix),
\beq
	\hat \y_m = \h_j^T\h_i/\left\|\h_i\right\|^2. \label{novel_feature}
\eeq
This equation implements a similarity-based inductive projection of the novel property to other items, where the similarity metric is precisely the Euclidean similarity of hidden representations of pairs of items (\figinductgen B). In essence, being told ``a pine has property \textit{x},'' the network will more readily project the novel property \textit{x} to those familiar items whose hidden representations are close to that of the pine.  

A parallel situation arises upon learning that a novel item possesses a familiar feature (e.g., ``a \textit{blick} can move,'' \figinductgen C). Encoding this knowledge requires new synaptic weights between the item and the hidden layer. Appropriate weights may be found through standard gradient descent learning of the item-to-hidden weights for this novel item, while holding the hidden-to-output weights fixed to preserve prior knowledge about features. The network can then inductively project other familiar properties to the novel item (e.g., "Does a blick have legs?") by simply generating a feature output vector in response to the novel item as input.  Under this scheme, a novel item $i$ with a familiar feature $m$ will be assigned another familiar feature $n$ through the equation (SI Appendix),
\beq
	\hat \y_n = \h_n^T\h_m/\left\|\h_m\right\|^2, \label{novel_item}
\eeq 
where the $\alpha^{th}$ component of $\h_n$ is $\h_n^\alpha= \uv^\alpha_n \sqrt{a_\alpha(t)}$.  $\h_n \in {\bf  R}^{N_2} $ can be thought of as the hidden representation of feature $n$ at developmental time $t$.  In parallel to \req{novel_feature}, this equation now implements similarity based inductive projection of familiar features to a novel item. In essence, being told ``a \textit{blick} can move,'' the network will more readily project other familiar features to a \textit{blick}, if those features have a similar internal representation as that of the feature move.

Thus the hidden layer of the deep network furnishes a common, semantic representational space into which both features and items can be placed.  When a novel feature $m$ is assigned to a familiar item $i$, that novel feature is placed close to the familiar item in the hidden layer, and so the network will inductively project this novel feature to other items close to $i$ in neural space.  In parallel, when a novel item $i$ is assigned a familiar feature $m$, that novel item is placed close to the familiar feature, and so the network will inductively project other features close to $m$ in neural space, onto the novel item.  

\par This principle of similarity based generalization encapsulated in Eqns. \ref{novel_feature} and \ref{novel_item}, when combined with the progressive differentiation of internal representations over developmental time as the network is exposed to hierarchically structured data, as illustrated in \figprogdiff B, then naturally explains the shift in patterns of inductive projection from broad to specific across development, as shown in \figinductgen E.  For example, consider specifically the inductive projection of a novel feature to familiar items (\figinductgen AB).  Earlier (later) in developmental time, neural representations of all items are more similar to (different from) each other, and so the network similarity based inductive projection will extend the novel feature to many (fewer) items, thereby exhibiting progressively narrower patterns of inductive projection that respect the hierarchical similarity structure of the environment (\figinductgen E).  Thus remarkably, even a deep linear network can provably exhibit the same broad to specific changes in patterns of inductive projection that are empirically observed in many works \cite{Carey1985,Carey2011,Keil1991,Murphy2002}.

\section{Linking Behavior and Neural Representations}
Compared to previous models which have primarily made behavioral predictions, our theory has a clear neural interpretation. Here we discuss implications for the neural basis of semantic cognition.

\subsection{Similarity structure is an invariant of optimal learning}

An influential method for probing neural codes for semantic knowledge in empirical measurements of neural activity is the representational similarity approach (RSA) \cite{Edelman1998,Laakso2000,Esteky2011,Kriegeskorte2013}, which examines the similarity structure of neural population vectors in response to different stimuli. This technique has identified rich structure in high level visual cortices, where, for instance, inanimate objects are differentiated from animate objects \cite{Carlson2014,Carlson2013,Giordano2013,Liu2013,Connolly2012}. Strikingly, studies have found remarkable constancy between neural similarity structures across human subjects, and even between humans and monkeys \cite{Mur2013,Esteky2011}. This highly conserved similarity structure emerges despite considerable variability in neural activity patterns across subjects \cite{Haxby2001,Shinkareva2012}. Indeed, exploiting similarity structure enables more effective across-subject decoding of fMRI data relative to transferring a decoder based on careful anatomical alignment \cite{Raizada2012}. Why is representational similarity conserved, both across individuals and species, despite highly variable tuning of individual neurons and anatomical differences? 

Remarkably, we show that two networks trained in the same environment {\it must} have {\it identical} representational similarity matrices despite having detailed differences in neural tuning patterns, {\it provided} that the learning process is optimal, in the sense that it yields the smallest norm weights that solve the task (see SI Appendix for a derivation).  One way to get close to the optimal manifold of synaptic weights of smallest norm after learning, is to start learning from small random initial weights.  We show in \figlearninginvar AB that two networks, each starting from different sets of small 
\begin{figure}
\begin{center}
\includegraphics[width=3.0in]{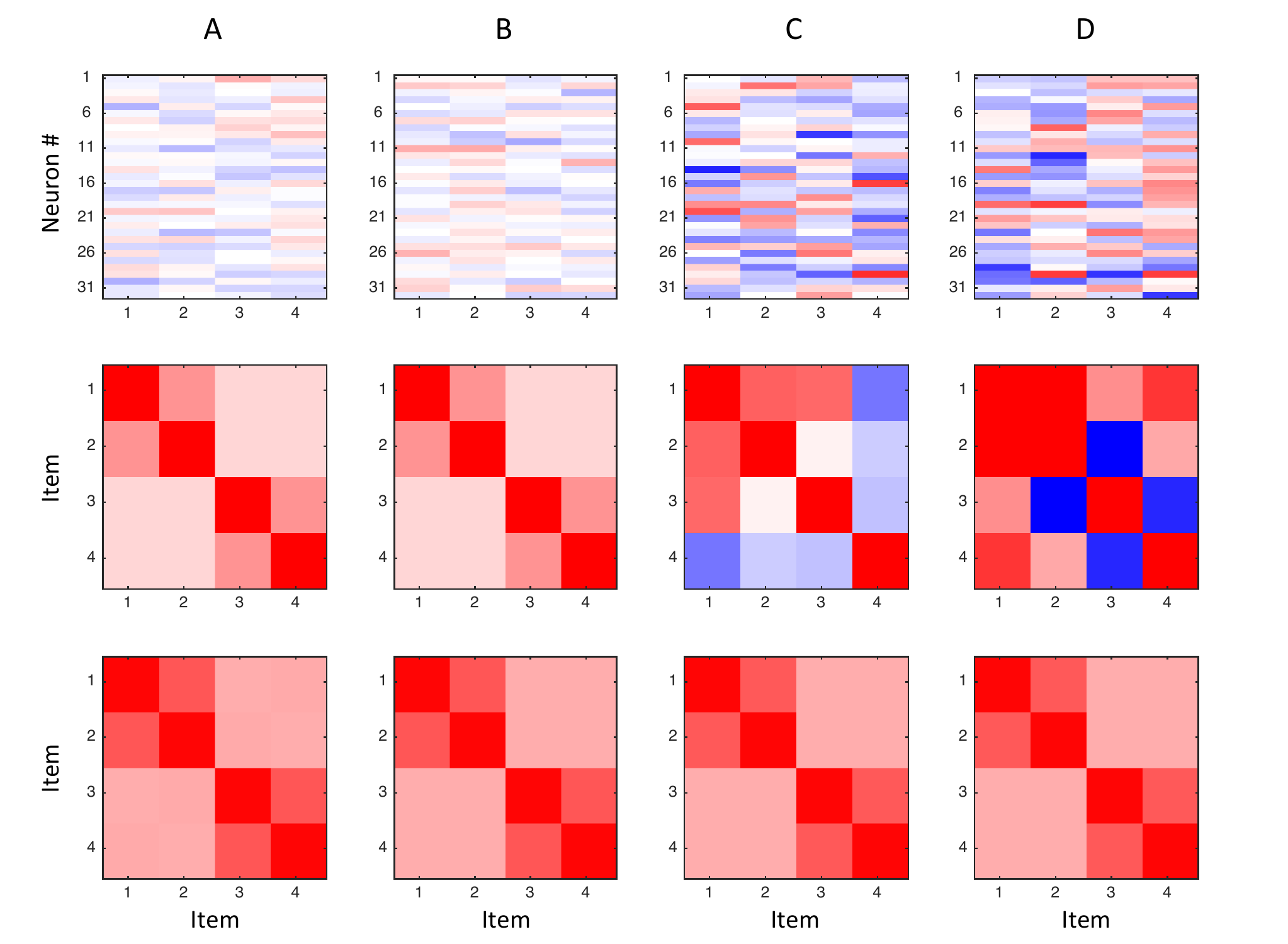}
\end{center}
\caption{Neural representations and invariants of learning. Columns A-B depict two networks trained from small norm random weights. Columns C-D depict two networks trained from large norm random weights. Top row: Neural tuning curves $h_i$ at the end of learning. Neurons show mixed selectivity tuning, and individual tuning curves are different for different trained networks. Middle row: Representational similarity matrix $\sh$. Bottom row: Behavioral similarity matrix $\shy$. For small-norm, but not large-norm weight initializations, representational similarity is conserved and behavioral similarity mirrors neural similarity.}
\label{figinvar}
\end{figure}
random initial weights, will after training learn very different internal representations (\figlearninginvar AB top row) but will have nearly identical representational similarity matrices (\figlearninginvar AB middle row).  Such a result is however, not obligatory.  Two networks starting from large random initial weights not only learn different internal representations, but also learn different representational similarity matrices (\figlearninginvar CD top and middle rows).  This pair of networks both learn the same composite input output map, but with suboptimal large-norm weights. Hence our theory, combined with the empirical finding that similarity structure is preserved across humans and species, may speculatively suggest that all these disparate neural circuits may be implementing an approximately optimal learning process in a common environment.

\subsection{When the brain mirrors behavior}
In addition to matching neural similarity patterns across subjects, experiments using fMRI and single unit responses have also documented a correspondence between neural similarity patterns and behavioral similarity patterns \cite{Kriegeskorte2013}. When does neural similarity mirror behavioral similarity? We show this correspondence again emerges only in optimal networks. 

In particular, denote by $\hat {\bf y}_i$ the behavioral output of the network in response to item $i$. These output patterns yield the behavioral similarity matrix $\shy_{ij}= \hat{\bf  y}_i^T \hat{\bf  y}_j$.  In contrast, the neural similarity matrix is $\sh_{ij} = {\bf  h}_i^T {\bf  h}_j$ where ${\bf  h}_i$ is the hidden representation of stimulus $i$.  We show in the SI Appendix that if the network learns optimal smallest norm weights, then these two similarity matrices obey the relation
\begin{equation}
\shy  = \left(\sh\right)^2.
\end{equation}
Moreover, we show the two matrices share the same singular vectors. Hence behavioral similarity patterns share the same structure as neural similarity patterns, but with each semantic distinction expressed more strongly (according to the square of its singular value) in behavior relative to the neural representation. %Intuitively, this greater distinction in behavior is due to the fact that half of the behavioral output relation is encoded in the output weights $\wb$, which do not influence the neural similarity of the hidden layer, as it depends only on $\wa$. 
While this precise mathematical relation is yet to be tested in detail, some evidence points to this greater category separation in behavior \cite{Mur2013}. 

Given that optimal learning is a prerequisite for neural similarity mirroring behavioral similarity, as in the previous section, there is a match between the two for pairs of networks trained from small random initial weights (\figlearninginvar AB middle and bottom rows), but not for pairs of networks trained from large random initial weights (\figlearninginvar CD middle and bottom rows).  Thus again, speculatively, our theory suggests that the experimental observation of a link between behavioral and neural similarity may in fact indicate that learning in the brain is finding optimal network solutions that efficiently implement the requisite transformations with minimal synaptic strengths.

\vspace{-2mm}
\section{Discussion}

In summary, the main contribution of our work is the analysis of a simple model, namely a deep linear neural network, that can, surprisingly, qualitatively capture a diverse array of phenomena in semantic development and cognition.  Our exact analytical solutions of nonlinear learning phenomena in this model yield conceptual insights into why such phenomena also occur in more complex nonlinear networks \cite{Rogers2004,Rumelhart1993,McClelland2005,Plunkett1992,Quinn1997,Colunga2005,McMurray2012,Blouw2016} trained to solve semantic tasks.  In particular, we find that the hierarchical differentiation of internal representations in a deep, but not a shallow, network (\figprogdiff) is an inevitable consequence of the fact that singular values of the input-output correlation matrix drive the timing of rapid developmental transitions (\figsvd~and Eqns.~\req{adynam} and \req{tauovs_deep}), and hierarchically structured data contains a hierarchy of singular values (\fighiersvd).  In turn, semantic illusions can be highly prevalent between these rapid transitions simply because global optimality in predicting all features of all items necessitates sacrificing correctness in predicting some features of some items (\figilluscorrs).  And finally, this hierarchical differentiation of concepts is intimately tied to the progressive sharpening of inductive generalizations made by the network (\figinductgen). 

The encoding of knowledge in the neural network after learning also reveals precise mathematical definitions of several aspects of semantic cognition.  Basically, the synaptic weights of the neural network extract from the statistical structure of the environment a set of paired object analyzers and feature synthesizers associated with every categorical distinction.  The bootstrapped, simultaneous  learning of each pair solves the apparent Gordian knot of knowing both which items belong to a category, and which features are important for that category: the object analyzers determine category membership, while the feature synthesizers determine feature importance, and the set of extracted categories are {\it uniquely} determined by the statistics of the environment.  Moreover, by defining the typicality of an item for a category as the strength of that item in the category's object analyzer, we can prove that typical items must enjoy enhanced performance in semantic tasks relative to atypical items (Eq.~\req{typicality_feat}). Also, by defining the category prototype to be the associated feature synthesizer, we can prove that the most typical items for a category are those that have the most {\it extremal} projections onto the category prototype (\figtypgeom~and Eq.~\ref{v_typ} ).  Finally, by defining the coherence of a category to be the associated singular value, we can prove that more coherent categories can be learned more easily and rapidly (\figcategcoh) and explain how changes in the statistical structure of the environment determine what level of a category hierarchy is the most basic or important (\figbasiclvl).    All our definitions of typicality, prototypes and category coherence are broadly consistent with intuitions articulated in a wealth of psychology literature, but our definitions imbue these intuitions with enough mathematical precision to prove theorems connecting them to aspects of category learnability, learning speed and semantic task performance in a neural network model.  

More generally, beyond categorical structure, our analysis provides a principled framework for explaining how the statistical structure of diverse structural forms associated with different probabilistic graphical models gradually become encoded in the weights of a neural network (\figstructform). 
Also, with regards to neural representation, our theory reveals that across different networks trained to solve a task, while there may be no correspondence at the level of single neurons, the similarity structure of internal representations of any two networks will both be {\it identical} to each other, and closely related to the similarity structure of behavior, {\it provided} both networks solve the task optimally, with the smallest possible synaptic weights.  Neither correspondence is obligatory, in that both need not hold for suboptimal networks (\figlearninginvar).  This result suggests, but by no means proves, that the neural and behavioral alignment of similarity structure in human and monkey IT may be a consequence of each circuit finding optimal solutions to similar tasks. 

While our simple neural network surprisingly captures this diversity of semantic phenomena in a mathematically tractable manner, because of its linearity, the phenomena it can capture still barely scratch the surface of semantic cognition. Some fundamental semantic phenomena that require complex nonlinear processing and memory include context dependent computations, dementia in damaged networks, theory of mind, the deduction of causal structure, and the binding of items to roles in events and situations. While it is inevitably the case that biological neural circuits exhibit all of these phenomena, it is not clear how our current generation of artificial nonlinear neural networks can recapitulate all of them.  However, we hope that a deeper mathematical understanding of even the simple network presented here can serve as a springboard for the theoretical analysis of more complex neural circuits, which in turn may eventually shed much needed light on how the higher level computations of the mind can emerge from the biological wetware of the brain.

\begin{acknowledgments}
S.G. thanks the Burroughs-Wellcome, Sloan, McKnight, James S. McDonnell and Simons Foundations for support.  J.L.M. was supported by AFOSR. A.S. was supported by Swartz, NDSEG, \& MBC Fellowships.  We thank Juan Gao and Madhu Advani for useful discussions.
\end{acknowledgments}
%\vspace{-.4in}
\bibliographystyle{unsrt}
\bibliography{semantic_cog}

\end{article}

\end{document}

\title{Supplementary Material}

\author{Andrew M. Saxe\affil{1}{University of Oxford, Oxford, UK},
 	James L. McClelland\affil{2}{Stanford University, Stanford, CA},    
	\and
	Surya Ganguli\affil{2}{}\affil{3}{Google Brain, Mountain View, CA}}

\contributor{}

\maketitle

\begin{article}
\begin{abstract}

\end{abstract}

\section{Acquiring Knowledge}

We consider the setting where we are given a set of $P$ examples $\left\{\x^i,\y^i\right\}, i=1,\ldots, P$, where the input vector $\x^i$ identifies item $i$, and the output vector $\y^i$ is a set of features to be associated to this item. The network, defined by the weight matrices $\wa,\wb$ in the case of the deep network (or $\ws$ in the case of the shallow network), computes its output as
\begin{equation}
	\hat \y = \wb \wa \x \label{network_map} 
\end{equation}
(or $\hat \y=\ws \x$ for the shallow network). Training proceeds through stochastic gradient descent on the squared error
\begin{equation}
SSE(\wa,\wb)=\frac{1}{2}\left\| \y^i - \hat \y^i \right\|^2, \label{error}
\end{equation}
with learning rate $\lambda$, yielding the updates
\begin{eqnarray*}
	\Delta \wa & = &  - \lambda \frac{\partial}{\partial \wa}SSE(\wa,\wb) \\
	\Delta \wb & = &  - \lambda \frac{\partial}{\partial \wb}SSE(\wa,\wb).
\end{eqnarray*}
While the structure of the error surface in such deep linear networks is known \cite{Baldi1989}, our focus here is on the dynamics of the learning process. Substituting Eqn.\,\req{network_map} into Eqn.\,\req{error} and taking derivatives yields the update rules specified in \wdeltabp\, of the main text,
\begin{eqnarray*}
	\Delta \wa & = & \lambda \wb^T \left( \y^i - \hat \y^i  \right) \x^{i T}\label{w_delta_bp}, \\
	\Delta \wb  & = & \lambda  \left( \y^i - \hat \y^i \right) \h^{i T},
\end{eqnarray*}
where $\h^i=\wa \x^i$ is the hidden layer activity for example $i$.

This update is identical to that produced by the standard backpropagation algorithm, as can be seen by noting that the error ${\bf e}^i=  \y^i - \hat \y^i$, and the backpropagated delta signal ${\bf \delta}^i=\wb^T{\bf e}$, such that these update equations can be rewritten as 
\begin{eqnarray*}
	\Delta \wa & = & \lambda {\bf \delta}^i \x^{i T}\label{w_delta_standard_bp}, \\
	\Delta \wb  & = & \lambda  {\bf e}^i \h^{i T}. 
\end{eqnarray*}

We now derive the average weight change under these updates over the course of an epoch, when learning is gradual. We assume that all inputs $i=1,\cdots,P$ are presented (possibly in random order), with updates applied after each. In the updates above, the weights change on each stimulus presentation and hence are functions of $i$, which we denote as $\wa[i],\wb[i]$. Our goal is to recover equations describing the dynamics of the weights across epochs, which we denote as $\wa(t),\wb(t)$. Here, $t=1$ corresponds to viewing $P$ examples, $t=2$ corresponds to viewing $2P$ examples, and so on. In general throughout the main text and supplement we suppress this dependence for clarity where it is clear from context and simply write $\wa,\wb$. 

When learning is gradual ($\lambda \ll 1$), the weights change minimally on each given example and hence $\wa[i]\approx\wa(t)$ for all patterns in epoch $t$. The total weight change over an epoch is thus
\begin{eqnarray*}
	\Delta \wa(t)& = & \sum_{i=1}^P \lambda \wb[i]^T \left( \y^i - \hat \y^i  \right) \x^{i T}, \\
	& = & \sum_{i=i}^P \lambda \wb[i]^T \left( \y^i - \wb[i]\wa[i]\x^i  \right) \x^{i T}, \\
	& \approx & \sum_{i=1}^P \lambda \wb(t)^T \left( \y^i - \wb(t)\wa(t)\x^i  \right) \x^{i T}, \\
%	& \approx &   \lambda \wb(t)^T \sum_i\left( \y^i - \hat \y^i  \right) \x^{i T},\\
%	& = &   \lambda P \wb(t)^T E[ \y^i\x^{i T} - \hat \y^i   \x^{i T}],\\
	& = &   \lambda P \wb(t)^T (E[ \y\x^{ T}] - \wb(t)\wa(t)  E[\x \x^{ T}]),\\
	& = & \lambda P\wb^T \left( \sio - \wb(t) \wa(t) \si \right)\\
	\Delta \wb(t)& = & \sum_{i=1}^P \lambda \left( \y^i - \hat \y^i  \right) \h^{i T} \\
	%& = & \sum_{i=1}^P \lambda \left( \y^i - \hat \y^i  \right) \x^{i T}\wa[i]^T \\
	& = & \sum_{i=i}^P \lambda \left( \y^i - \wb[i]\wa[i]\x^i  \right) \x^{i T}\wa[i]^T , \\
	& \approx & \sum_{i=1}^P \lambda  \left( \y^i - \wb(t)\wa(t)\x^i  \right) \x^{i T}\wa(t)^T, \\
%	& \approx &   \lambda \wb(t)^T \sum_i\left( \y^i - \hat \y^i  \right) \x^{i T},\\
%	& = &   \lambda P \wb(t)^T E[ \y^i\x^{i T} - \hat \y^i   \x^{i T}],\\
	& = &   \lambda P (E[ \y\x^{ T}] - \wb(t)\wa(t)  E[\x \x^{ T}])\wa(t)^T,\\
	& = & \lambda P\left( \sio - \wb(t) \wa(t) \si \right)\wa(t)^T.
\end{eqnarray*}
where  
$\si \equiv  E[\x \x^T]$ is an $N_1 \times N_1$ input correlation matrix, and
$\sio \equiv  E[\y \x^T] \label{sio_def}$ is an $N_3 \times N_1$ input-output correlation matrix.
So long as $\lambda$ is small, we can take the continuum limit of this difference equation to obtain Eqns.\waavg \,of the main text,
\begin{eqnarray}
	\tau \ddt \wa & = & \wb^T \left( \sio - \wb \wa \si \right),  \label{wa_avg}\\
	\tau \ddt \wb & = &   \left( \sio - \wb \wa \si \right) \wa^T.\label{wb_avg}
\end{eqnarray}
where the time constant 

\begin{equation}
	\tau \equiv \frac{1}{P\lambda}. \label{tau_def}
\end{equation} 

In the above, the weights are now a function of a continuous parameter that with slight abuse of notation we also denote as $t$, such that as $t$ goes from $0$ to $1$ the network has seen $P$ examples.

\subsection{Explicit solutions from tabula rasa}

To solve for the dynamics of $\wa,\wb$ over time, we decompose the input-output correlations through the singular value decomposition (SVD),
\beq
	\sio = \ls \sv \rs^T = \sum_{\alpha=1}^{N_1} s_\alpha \uv^\alpha \vv^{\alpha T}, \label{eq_svd} \nonumber
\eeq
and then change variables to $\wao,\wbo$ where
\begin{eqnarray}
	\wa & = & \br\wao \rs^T, \label{cov_a} \\
	\wb &= & \ls \wbo\br^T, \label{cov_b}
\end{eqnarray}
and $\br$ is an arbitrary orthogonal matrix ($\br^T \br=I$).
These variables analyze the dynamics in the basis defined by the SVD. Substituting into Eqns.\,\req{wa_avg}-\req{wb_avg} and using the simplifying assumption $\si=I$ we have
\begin{eqnarray}
	\tau \ddt (\br\wao\rs^T) & = & \br\wbo^T\ls^T \left( \sio - \ls\wbo \wao \rs^T \si \right),  \nonumber \\
	\tau \ddt \wao & = & \wbo^T\ls^T \left( \ls\sv\rs^T - \ls\wbo \wao \rs^T  \right)\rs,\nonumber\\
	 & = &  \wbo^T \left( \sv - \wbo \wao   \right),\label{wa_trans}\\
	\tau \ddt (\ls\wbo\br^T) & = &   \left( \sio - \ls\wbo \wao \rs^T \si \right) \rs^T\wao^T\br^T\nonumber\\
	\tau \ddt \wbo & = &   \ls^T\left( \ls \sv \rs^T - \ls\wbo \wao \rs^T  \right) \rs\wao^T \nonumber\\
	 & = &   \left(  \sv  - \wbo \wao   \right) \wao^T \label{wb_trans}
\end{eqnarray}
where we have made use of the orthogonality of the SVD bases, i.e., $\rs^T\rs=I$ and $\ls^T\ls=I$. Importantly, the change of variables is applied \textit{after} deriving the gradient descent update equations in the untransformed coordinate system. Gradient descent is not invariant to reparametrization and so performing this change of variables \textit{before} would correspond to analyzing potentially different dynamics. 

Equations \req{wa_trans}-\req{wb_trans} have a simplified form because $\sv$ is a diagonal matrix. Hence if $\wao$ and $\wbo$ are also diagonal, the dynamics decouple into $N_1$ independent systems. We study the dynamics in this decoupled regime where $\wao(0)$ and $\wbo(0)$ are diagonal. Off-diagonal elements represent coupling between different modes of the SVD, and decay to zero under the dynamics. Hence the decoupled solutions we find also provide good approximations to the full solution when $\wao(0)$ and $\wbo(0)$ are initialized with small random weights, as shown through simulation (red lines in \figsvd C of the main text).

In particular, let $c_\alpha = \wao_{\alpha\alpha}$ and $d_\alpha = \wbo_{\alpha\alpha}$ be the $\alpha^{th}$ diagonal element of the first and second matrices, encoding the strength of mode $\alpha$ transmitted by the input-to-hidden and hidden-to-output weights respectively. We have the scalar dynamics
\begin{eqnarray*}
	\tau\ddt c_\alpha & = & d_\alpha(s_\alpha - c_\alpha d_\alpha) \\
	\tau\ddt d_\alpha & = & c_\alpha(s_\alpha - c_\alpha d_\alpha)
\end{eqnarray*}
for $\alpha=1,\cdots,N_1$. In general, $c_\alpha$ can differ from $d_\alpha$, but if weights are initialized to small initial values, these will be roughly equal. We therefore study \textit{balanced} solutions where $c_\alpha=d_\alpha$. In particular, we will track the overall strength of a particular mode with the single scalar $a_\alpha=c_\alpha d_\alpha$, with dynamics
\begin{eqnarray}
	\tau \ddt a_\alpha & = &   c_\alpha \left(\tau\ddt d_\alpha\right) +  \tau d_\alpha \left(\tau\ddt c_\alpha\right) \nonumber\\
	& = &  c_\alpha c_\alpha(s_\alpha - c_\alpha d_\alpha) +\tau d_\alpha d_\alpha(s_\alpha - c_\alpha d_\alpha)\nonumber \\
	& = & 2 a_\alpha(s_\alpha-a_\alpha). \nonumber
\end{eqnarray}
This is a separable differential equation which can be integrated to yield (here we suppress the dependence on $\alpha$ for clarity),
\begin{eqnarray}
	t & = & \frac{\tau}{2} \int_{a^0}^{a^f} \frac{da}{a(s-a)}  =  \frac{\tau}{2s}\ln \frac{a^f(s-a^0)}{a^0(s-a^f)}\label{timescale}
\end{eqnarray}
where $t$ is the time required to travel from an initial strength $a(0)=a^0$ to a final strength $a(t)=a^f$. 

The entire time course of learning can be found by solving for $a_f$, yielding Eqn.\adynam \,of the main text,
\begin{equation*}
	a_\alpha(t)=\frac{s_\alpha e^{2s_\alpha t/\tau}}{e^{2s_\alpha t/\tau}-1+s_\alpha/a_\alpha^0}. \label{adynam}
\end{equation*}

Next, we undo the change of variables to recover the full solution. Define the time-dependent diagonal matrix $\at$ to have diagonal elements $(\at)_{\alpha\alpha}=a_\alpha(t)$. Then by the definition of $a_\alpha,c_\alpha,$ and $d_\alpha$, we have $\at=\wbo(t)\wao(t)$.
Inverting the change of variables in Eqns.~\req{cov_a}-\req{cov_b}, we recover Eqn.\timecourse of the main text, the overall input-output map of the network:
\begin{eqnarray*}
	\wb(t)\wa(t) =  \ls \wbo(t)\wao(t) \rs^T=  \ls \at \rs^T.
\end{eqnarray*}
This solution is not fully general, but rather provides a good account of the dynamics of learning in the network in a particular regime. To summarize our assumptions, the solution is applicable in the \textit{gradual learning} regime ($\lambda \ll 1$), when initial mode strengths in each layer are roughly \textit{balanced} ($c_\alpha=d_\alpha$), and approximately \textit{decoupled} (off diagonal elements of $\wao,\wbo \ll 1$). These latter two conditions hold approximately when weights are initialized with small random values, and hence we call this solution the solution from \textit{tabula rasa}. Notably, these solutions do not describe the dynamics if substantial knowledge is already embedded in the network when learning commences. When substantial prior knowledge is present, learning can have very different dynamics corresponding to unequal initial information in each layer ($c_\alpha \neq d_\alpha$) and/or strong coupling between modes (large off-diagonal elements in $\wao,\wbo$).

How small must $\lambda$ be to count as gradual learning? The requirement on $\lambda$ is that the fastest dynamical timescale in \req{wa_avg}-\req{wb_avg} is much longer than 1, which is the timescale of a single learning epoch. The fastest  timescale arises from the largest singular value $s_1$ and is $O(\tau/s_1)$ (cf Eqn.~\req{timescale}). Hence the requirement $\tau/s_1 \gg 1$ and the definition of $\tau$ yields the condition
\begin{equation*}
	\lambda \ll \frac{1}{s_1P}.
\end{equation*}
Hence stronger structure, as measured by the SVD, or more training samples, necessitates a smaller learning rate.

The dynamics permit an explicit curve for the sum squared error over the course of learning. This is 
\begin{eqnarray*}
	SSE(t)&=&\frac{1}{2}\sum_{i=1}^P||\y^i-\hat \y^i||_2^2 \nonumber\\ 
	      %    & = & \frac{1}{2}\sum_{i=1}^P{\y^i}^T\y^i-2{\y^i}^T\hat \y^i + \hat {\y}^{i^T }\hat \y^i \\ 
	& = & \frac{1}{2}\tr \sum_{i=1}^P\y^i{\y^i}^T-2\hat \y^i{\y^i}^T + \hat \y^i \hat {\y}^{i^T } \nonumber\\ 
	%& = & \frac{1}{2} \tr YY^T - \tr \hat YY^T + \frac{1}{2}\tr \hat Y \hat Y^T \\
	& = &\frac{P}{2} \tr \so-P\tr \sio \Wtot^T + \frac{P}{2}\tr \Wtot \si \Wtot^T\nonumber \\
%	\frac{P}{2} \tr \so-P\tr \rs\sv\ls^T\ls \at \rs^T + \frac{P}{2}\tr (\at^2) \\
	& = & \frac{P}{2} \tr \so-P\tr\sv\at + \frac{P}{2}\tr \at^2\nonumber \\
	& = & \frac{P}{2} \tr \so - P \tr\left[\left(\sv - \frac{1}{2}\at\right)\at\right]. 
\end{eqnarray*} 
%This last expression is a sum over independent contributions $(SSE(t) = \sum_\alpha E_\alpha(t))$ from each mode, where the error arising from mode $\alpha$ is 
%\begin{eqnarray}
%	E_\alpha(t) & = & \frac{P}{2} \left[ \gamma_\alpha - 2(s_\alpha - a_\alpha(t)/2)a_\alpha(t)\right].
%\end{eqnarray}
%Here $\gamma_\alpha$ is the associated eigenvalue of the output correlation matrix $\so$. 
Early in learning, $\at\approx 0$ and the error is proportional to $\tr\so$, the variance in the output. Late in learning, $\at\approx \sv$ and the error is proportional to $\tr\so - \tr\sv^2$, the output variance which cannot be explained by a linear model.

In the \textit{tabula rasa} regime, the individual weight matrices are given by 
\begin{eqnarray*}
\wa(t) &  = & {\br} \wao\rs^T = {\br} \sqrt{\at}\rs^T, \label{internalrep1} \\
 \wb(t)  & = &  \ls \wbo\br^T=\ls \sqrt{\at}\br^T, \label{internalrep}
\end{eqnarray*}
due to the fact that $c_\alpha=d_\alpha=\sqrt{a_\alpha}$. 

The full space of weights implementing the same input-output map is  
\begin{eqnarray*}
\wa(t) &  = & \bq \sqrt{\at}\rs^T,\\
 \wb(t)  & = &  \ls  \sqrt{\at}\bq^{-1} 
\end{eqnarray*}
for any invertible matrix $\bq$.

\subsubsection{Shallow network}

Analogous solutions may be found for the shallow network. In particular, the gradient of the sum of square error \begin{equation*}
SSE(\ws)=\frac{1}{2}\left\| \y^i - \hat \y^i \right\|^2, \label{error_shallow}
\end{equation*}
yields the update
\begin{equation*}
	\Delta \ws = \lambda(\y^i-\hat \y^i)\x^{i T}.
\end{equation*}
Averaging as before over an epoch yields the dynamics
\begin{equation*}
	\tau \ddt \ws = \sio - \ws \si,
\end{equation*}
a simple linear differential equation which may be solved explicitly. To make the solution readily comparable to the deep network dynamics, we change variables to $\ws=\ls \wso \rs^T$, 
\begin{eqnarray*}
	\tau \ddt (\ls \wso \rs^T) & = & \sio - \ls \wso \rs^T \si \\
	\tau \ddt \wso & = & \sv - \wso.
\end{eqnarray*}
Defining $\wso_{\alpha\alpha}=b_\alpha$ and assuming decoupled initial conditions gives the scalar dynamics 
\begin{eqnarray*}
	\tau \ddt b_\alpha = s_\alpha - b_\alpha.
\end{eqnarray*}
Integrating this simple separable differential equation yields 
\begin{equation}
	t =  \tau\ln\frac{s_\alpha - b^0_\alpha}{s_\alpha-b^f_\alpha} \label{s_timescale}
\end{equation}
which can be inverted to find the full time course
\beq
	b_\alpha(t) = s_\alpha \left(1-e^{-t/ \tau}\right) + b_\alpha^0e^{-t/\tau}. \label{bdynam} \nonumber
\eeq
Undoing the change of variables yields the weight trajectory 
\beq
	\ws = \ls \bt \rs^T \nonumber
\eeq
where $\bt$ is a diagonal matrix with elements $(\bt)_{\alpha\alpha}=b_\alpha(t).$
\subsection{Simulation details for solutions from tabula rasa}

The simulation results shown in \figsvd~are for a minimal hand-crafted hierarchical dataset with $N_3=7$ features, $N_2=16$ hidden units, and $N_1=P=4$ items. Inputs were encoded with one-hot vectors. The input-output correlations are
\begin{eqnarray*}
    \sio &=& 0.7P\begin{bmatrix}
    1 & 1 & 1 & 1 \\
    1 & 1 & 0 & 0 \\
    0 & 0 & 1 & 1 \\
    1 & 0 & 0 & 0 \\
    0 & 1 & 0 & 0 \\
    0 & 0 & 1 & 0 \\
    0 & 0 & 0 & 1
    \end{bmatrix}\\
    \si &=& {\bf I}.
\end{eqnarray*}
We used $\tau=1$, $\lambda=\frac{1}{P}$, and $a_0=0.0001$.

\subsection{Rapid stage-like transitions due to depth}

To understand the time required to learn a particular mode, we calculate the time $t$ necessary for the learning process to move from an initial state with little knowledge, $a(0)=\epsilon$ for some small $\epsilon \ll 1$, to a state which has reached within $\epsilon$ of its final asymptote, $a(t_f)=s - \epsilon$. It is necessary to introduce this cutoff parameter $\epsilon$ because, first, deep networks initialized with weights exactly equal to zero have no dynamics, and second, because both shallow and deep networks do not reach their asymptotic values in finite time. Therefore we consider networks initialized a small distance away from zero, and consider learning to be complete when they arrive within a small distance of the correct answer.  For the deep network, substituting these initial and final conditions into Eqn.~\req{timescale} yields
\begin{eqnarray*}
%t & = &  \frac{\tau}{2s}\ln \frac{a^f(s-a^0)}{a^0(s-a^f)}\\
t	&  = &  \frac{\tau}{2s}\ln \frac{(s-\epsilon)^2}{\epsilon^2} \\
	& \approx & \frac{\tau}{s}\ln \frac{s}{\epsilon} 
\end{eqnarray*}
for small $\epsilon$.

For the shallow network, by contrast, substituting into Eqn.~\req{s_timescale} yields
\begin{eqnarray*}
t & = &  \tau\ln\frac{s- \epsilon}{\epsilon} \\
 & \approx & \tau\ln\frac{s}{\epsilon}
\end{eqnarray*}
for small $\epsilon$. Hence these networks exhibit fundamentally different learning timescales, due to the $1/s$ term in the deep network, which strongly orders the learning times of different modes by their singular value size.

Beyond this difference in learning timescale, there is a qualitative change in the shape of the learning trajectory. Deep networks exhibit sigmoidal learning trajectories for each mode, while shallow networks undergo simple exponential approach. The sigmoidal trajectories in deep networks give a quasi-stage-like character to the learning dynamics: for much of the total learning time, progress is very slow; then in a brief transition period, performance rapidly improves to near its final value. How does the length of the transitional period compare to the total learning time? We define the transitional period as the time required to go from a strength $a(t_{s})=\epsilon$ to within a small distance of the asymptote, $a(t_f)=s-\epsilon$, as before. Here $t_s$ is the time marking the start of the transition period and $t_f$ is the time marking the end. Then we introduce a new cutoff $\epsilon_0<\epsilon$ for the starting strength of the mode, $a(0)=\epsilon_0$. The length of the transition period $t_{trans}=t_f-t_s$ is
\begin{eqnarray*}
	t_{trans}& = &\frac{\tau}{2s}\ln \frac{(s-\epsilon)^2}{\epsilon^2},
\end{eqnarray*}
while the total learning time $t_{tot}=t_f$ starting from the mode strength $\epsilon_0$ is 
\begin{eqnarray*}
	t_{tot}& = & \frac{\tau}{2s}\ln \frac{(s-\epsilon)(s-\epsilon_0)}{\epsilon_0\epsilon}.
\end{eqnarray*}
Hence for a fixed $\epsilon$ defining the transition period, the total training time increases as the initial strength on a mode $\epsilon_0$ decreases toward zero. In the limit $\epsilon_0\rightarrow0$, the ratio of the length of time in the transition period to the total training time is
\begin{equation*}
	\lim_{\epsilon_0\rightarrow0} t_{trans}/t_{tot}  =  0,
\end{equation*}
such that the duration of the transition is exceptionally brief relative to the total training time. Hence deep networks can exhibit stage-like transitions.

By contrast, for the shallow network,
\begin{eqnarray*}
t_{trans}& = &  \tau\ln\frac{s - \epsilon}{\epsilon} \\
t_{tot}& = &  \tau\ln\frac{s - \epsilon_0}{\epsilon} 
\end{eqnarray*}
and the ratio limits to $t_{trans}/t_{tot} = 1$
for fixed small $\epsilon$ and $\epsilon_0\rightarrow 0$, indicating that the transition period is as long as the total training time and transitions are not stage-like.

\subsection{Progressive differentiation of hierarchical structure}
In this section we introduce a hierarchical probabilistic generative model of items and their attributes that, when sampled, produces a dataset that can be supplied to our simple linear network. Using this, we will be able to explicitly link hierarchical taxonomies to the dynamics of learning in our network. We show that our network will exhibit progressive differentiation with respect to any of the underlying hierarchical taxonomies allowed by our generative model.

A key result from the explicit solutions is that the time scale of learning of each input-output mode $\alpha$ of the correlation matrix $\sio$ is inversely proportional to the correlation strength $s_\alpha$ (i.e. singular value) of the mode. It is on this time scale, $O(\tau/s_\alpha)$, that the network learns to project perceptual representations onto internal representations using the right singular vector $\vv^\alpha$ of $\sio$, and then expand this component of the internal representation into a contribution to the predicted feature output vector given by $\uv^\alpha$, a left singular vector of $\sio$.

To understand the time course of learning of hierarchical structure, we analyze a simple generative model proposed in \cite{Kemp2004} of hierarchical data  $\{\x^\mu,\y^\mu\}$, and compute for this model the statistical properties $(s_\alpha, \uv^\alpha, \vv^\alpha)$ which drive learning.

\subsubsection{Hierarchical feature vectors from a branching diffusion process}
We first address the output data $\y^\mu, \mu = 1,\ldots,P$. Each $\y^\mu$ is an $N_3$-dimensional feature vector where each feature $i$ in example $\mu$ takes the value $\y_i^\mu=\pm1$. The value of each feature $i$ across all examples arises from a branching diffusion process occurring on a tree, as depicted in \sfigbranchingdiffusion. Each feature $i$ undergoes its own diffusion process on the tree, \textit{independent} of any other feature $j$. This entire process, described below, yields a hierarchical structure on the set of examples $\mu=1,\ldots,P$, which are in one-to-one correspondence with the leaves of the tree.

\begin{figure}
\begin{center}
\includegraphics[width=3.4in]{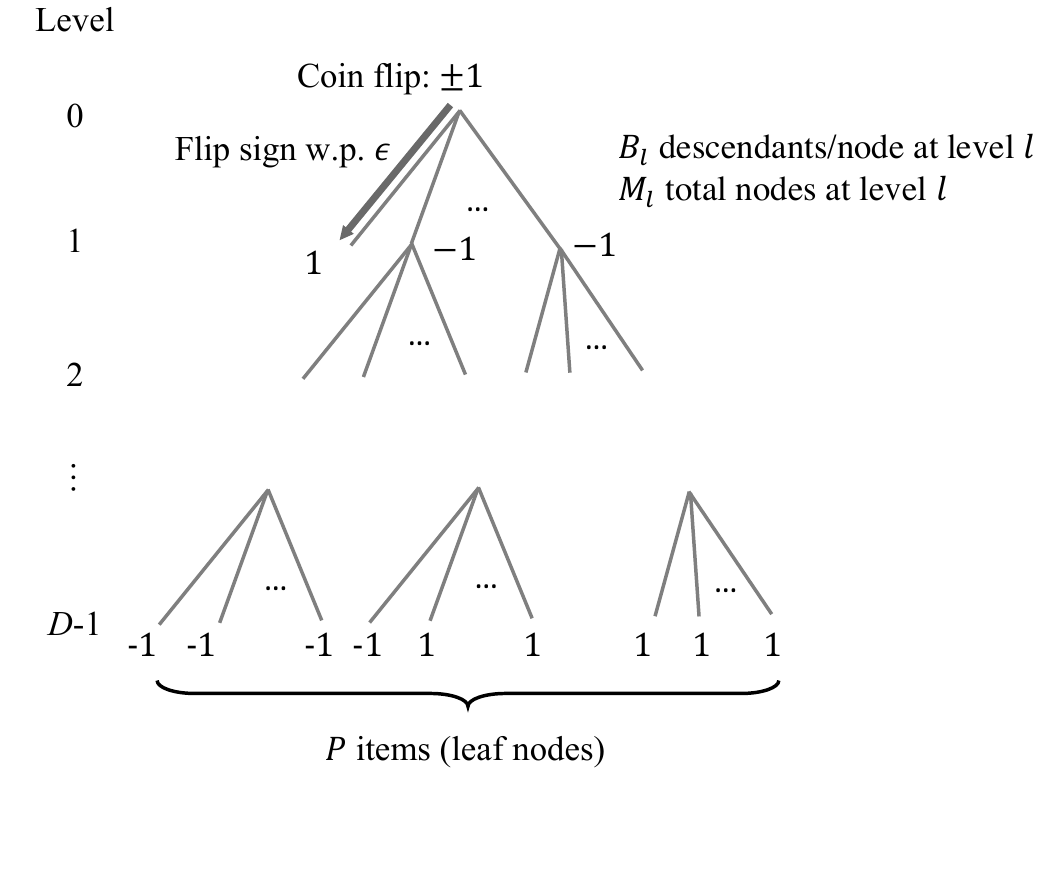}
\end{center}

\caption{Generating hierarchically structured data through a branching diffusion process. To generate a feature, an initial binary value is determined through a coin flip at the top of the hierarchy. The sign of this value flips with a small probability along each link in the tree. At the bottom, this yields the value of one feature across items. Many features can be generated by repeatedly sampling from this process independently. The $\pm 1$ values depicted are one possible sampling. }
\label{fig:branching_diffusion_process}
\end{figure}

The tree has a fixed topology, with $D$ levels indexed by $l=0,\ldots,D-1$, with $M_l$ total nodes at level $l$. We take for simplicity a regular branching structure, so that every node at level $l$ has exactly $B_l$ descendants. Thus $M_l=M_0\Pi^{l-1}_{k=0}B_l$. The tree has a single root node at the top ($M_0=1$), and again $P$ leaves at the bottom, one per example in the dataset ($M_{D-1}=P$).

Given a single feature component $i$, its value across $P$ examples is determined as follows. First draw a random variable $\eta^{(0)}$ associated with the root node at the top of the tree. The variable $\eta^{(0)}$ takes the values $\pm 1$ with equal probability $\frac{1}{2}$. Next, for each of the $B_0$ descendants below the root node at level 1, pick a random variable $\eta^{(1)}_i$, for $i=1,\ldots,B_0$. This variable $\eta^{(1)}_i$ takes the value $\eta^{(0)}$ with probability $1-\epsilon$ and $-\eta^{(0)}$ with probability $\epsilon$. The process continues down the tree: each of $B_{l-1}$ nodes at level $l$ with a common ancestor at level $l-1$ is assigned its ancestor's value with probability $1-\epsilon$, or is assigned the negative of its ancestor's value with probability $\epsilon$. Thus the original feature value at the root, $\eta^{(0)}$, diffuses down the tree with a small probability $\epsilon$ of changing at each level along any path to a leaf. The final values at the $P$ leaves constitute the feature values $\y_i^\mu$ for $\mu=1,\ldots,P$. This process is repeated independently for all feature components $i$.

In order to understand the dimensions of variation in the feature vectors, we would like to first compute the inner product, or overlap, between two example feature vectors. This inner product, normalized by the number of features $N_3$, has a well-defined limit as $N_3\rightarrow \infty$. Furthermore, due to the hierarchical diffusive process which generates the data, the normalized inner product only depends on the level of the tree at which the first common ancestor of the two leaves associated with the two examples arises. Therefore we can make the definition
\beq
	q_k=\frac{1}{N_3}\sum^{N_3}_{i=1}\y_i^{\mu_1}\y_i^{\mu_2}, \nonumber
\eeq
where again, the first common ancestor of leaves $\mu_1$ and $\mu_2$ arises at level $k$. It is the case that $1=q_{D-1}>q_{D-2}>\cdots>q_0>0$. Thus pairs of examples with a more recent common ancestor have stronger overlap than pairs of examples with a more distant common ancestor. These $D-1$ numbers $q_0,\ldots,q_{D-2}$, along with the number of nodes at each level $M_0,\ldots,M_{D-1}$, are the fundamental parameters of the hierarchical structure of the feature vectors; they determine the correlation matrix across examples, i.e.~the $P\times P$ matrix with elements
\beq
	{\bf \Sigma}_{\mu_1\mu_2}=\frac{1}{N_3}\sum^{N_3}_{i=1}\y_i^{\mu_1}\y_i^{\mu_2}, \label{hier_sig} 
\eeq
and hence its eigenvectors and eigenvalues, which drive network learning, as we shall see below.

It is possible to explicitly compute $q_k$ for the generative model described above. However, all that is really needed below is the property that $q_{D-2}>q_{D-1}>\cdots>q_0$. The explicit formula for $q_k$ is
\beq
	q_k=1-2\Omega \left(D-1-k, 2\epsilon(1-\epsilon)\right), \nonumber
\eeq
where $\Omega(N,P)$ is the probability that a sum of $N$ Bernoulli trials with probability $P$ of being 1 yields an odd number of $1$'s. It is clear that the overlap $q_k$ strictly decreases as the level $k$ of the last common ancestor decreases (i.e.~the distance up the tree to the last common ancestor increases).

\subsubsection{Input-output correlations for orthogonal inputs and hierarchical outputs}

We are interested in the singular values and vectors, $(s_\alpha, \uv^\alpha, \vv^\alpha)$ of $\sio$, since these drive the learning dynamics. We assume the $P$ output feature vectors are generated hierarchically as in the previous section, but then assume a localist representation in the input, so that there are $N_1=P$ input neurons and $\x_i^\mu=\delta_{\mu i}$. The input-output correlation matrix $\sio$ is then an $N_3 \times P$ matrix with elements $\sio_{i\mu}=\y_i^\mu$, with $i=1,\ldots,N_3$ indexing feature components, and $\mu=1,\ldots,P$ indexing examples. We note that
\beq
	(\sio)^T \sio = \rs \sv^T \sv \rs^T=N_3{\bf \Sigma}, \nonumber
\eeq
where ${\bf \Sigma}$, defined in \req{hier_sig}, is the correlation matrix across examples. From this we see that the eigenvectors of ${\bf \Sigma}$ are the same as the right singular vectors $\vv^\alpha$ of $\sio$, and if the associated eigenvalue of ${\bf \Sigma}$ is $\lambda_\alpha$, then the associated singular value of $\sio$ is $s_\alpha = \sqrt{N_3\lambda_\alpha}$. Thus finding the singular values $s_\alpha$ of $\sio$, which determine the time scales of learning, reduces to finding the eigenvalues $\lambda_\alpha$ of ${\bf \Sigma}$.

We note that the localist assumption $\x_i^\mu=\delta_{\mu i}$ is not necessary. We could have instead assumed an orthogonal distributed representation in which the vectors $\x^\mu$ form an orthonormal basis $O$ for the space of input-layer activity patterns. This would yield the modification $\sio \rightarrow \sio {\bf O}^T$, which would not change the singular values $s_\alpha$ at all, but would simply rotate the singular vectors, $\vv^\alpha$. Thus distributed orthogonal perceptual representations and localist representations yield exactly the same time course of learning. For simplicity, we focus here on localist input representations. 

We now find the eigenvalues $\lambda_\alpha$ and eigenvectors $\vv^\alpha$ of the correlation matrix across examples, ${\bf \Sigma}$ in \req{hier_sig}. This matrix has a hierarchical block structure, with diagonal elements $q_{D-1}=1$ embedded within blocks of elements of magnitude $q_{D-2}$ in turn embedded in blocks of magnitude $q_{D-3}$ and so on down to the outer-most blocks of magnitude $q_0 > 0$. This hierarchical block structure in turn endows the eigenvectors with a hierarchical structure.

To describe these eigenvectors we must first make some preliminary definitions. We can think of each $P$ dimensional eigenvector as a function on the $P$ leaves of the tree which generated the feature vectors $\y^\mu$, for $\mu=1,\ldots,P$. Many of these eigenvectors will take constant values across subsets of leaves in a manner that respects the topology of the tree. To describe this phenomenon, let us define the notion of a level $l$ funtion $f(\mu)$ on the leaves as follows: first consider a function $g$ which takes $M_l$ values on the $M_l$ nodes at level $l$ of the tree. Each leaf $\mu$ of the tree at level $D-1$ has a unique ancestor $\nu(\mu)$ at level $l$; let the corresponding level $l$ function on the leaves induced by $g$ be $f(\mu)=g(\nu(\mu))$. This function is constant across all subsets of leaves which have the same ancestor at level $l$. Thus any level $l$ function cannot discriminate between examples that have a common ancestor which lives at any level $l'>l$ (i.e.~any level lower than $l$).

Now every eigenvector of ${\bf \Sigma}$ is a level $l$ function on the leaves of the tree for some $l$. Each level $l$ yields a degeneracy of eigenvectors, but the eigenvalue of any eigenvector depends only on its level $l$. The eigenvalue $\lambda_l$ associated with every level $l$ eigenvector is 
\beq
	\lambda_l \equiv P\left( \sum_{k=l}^{D-1} \frac{\Delta_l}{M_l} \right), \nonumber
\eeq
where $\Delta_l \equiv q_l - q_{l-1}$, with the caveat that $q_{-1} \equiv 0$. It is clear that $\lambda_l$ is a decreasing function of $l$. This immediately implies that finer scale distinctions among examples, which can only be made by level $l$ eigenvectors for larger $l$, will be learned later than coarse-grained distinctions among examples, which can be made by level $l$ eigenvectors with smaller $l$.

We now describe the level $l$ eigenvectors. They come in $M_{l-1}$ families, one family for each node at the higher level $l-1$ ($l=0$ is a special case--there is only one eigenvector at this level and it is a uniform mode that takes a constant value on all $P$ leaves). The family of level $l$ eigenvectors associated with a node $\nu$ at level $l-1$ takes nonzero values only on leaves which are descendants of $\nu$. They are induced by functions on the $B_{l-1}$ direct descendants of $\nu$ which sum to 0. There can only be $B_{l-1}-1$ such orthonormal eigenvectors, hence the degeneracy of all level $l$ eigenvectors is $M_{l-1}(B_{l-1}-1)$. Together, linear combinations of all these level $l$ eigenvectors can be used to assign different values to any two examples whose first common ancestor arises at level $l$ but not at any lower level $l'>l$. Thus level $l$ eigenvectors do not see any structure in the data at any level of granularity below level $l$ of the hierarchical tree which generated the data. Recall that these eigenvectors are precisely the input modes which project examples onto internal representations in the multilayer network. Importantly, this automatically implies that structure below level $l$ in the tree cannot arise in the internal representations of the network until after structure at level $l-1$ is learned.

We can now be quantitative about the time scale at which structure at level $l$ is learned. We first assume the branching factors $B_l$ are relatively large, so that to leading order, $\lambda_l=P\frac{\delta_l}{M_l}$. Then the singular values of $\sio$ at level $l$ are
\beq
	s_l=\sqrt{N \lambda_l}=\sqrt{NP\frac{\Delta_l}{M_l}}. \nonumber
\eeq
The time scale of learning structure at level $l$ is then
\beq
	\tau_l=\frac{\tau}{s_l}=\frac{1}{\lambda}\sqrt{\frac{P}{N}\frac{M_l}{\Delta_l}}, \nonumber
\eeq
where we have used the definition of $\tau$ in \req{tau_def}. The fastest time scale is $\tau_0$ since $M_0=1$ and the requirement that $\tau_0\gg 1$ yields the requirement $\lambda \ll \sqrt{P/N}$. If we simply choose $\lambda=\epsilon\sqrt{P/N}$ with $\epsilon \ll 1$, we obtain the final result
\beq
	\tau_l = \frac{1}{\epsilon}\sqrt{\frac{M_l}{\Delta_l}}. \nonumber
\eeq
Thus the time scale for learning structure at a level of granularity $l$ down the tree, for this choice of learning rate and generative model, is simply proportional to the square root of the number of ancestors at level $l$. For constant branching factor $B$, this time scale grows exponentially with $l$.

\subsection{Illusory correlations}
The dynamics of learning in deep but not shallow networks can cause them to exhibit illusory correlations during learning, where the prediction for a particular feature can be a U-shaped function of time. This phenomenon arises from the strong dependence of the learning dynamics on the singular value size, and the sigmoidal stage-like transitions in the deep network. 
In particular, a feature $m$ for item $i$ receives a contribution from each mode $\alpha$ of $a_\alpha(t) \uv_m^\alpha \vv_i^\alpha$. 
Looking at two successive modes $k$ and $k+1$, these will cause the network's estimate of the feature to increase and decrease respectively if $\uv_m^{k}\vv_i^{k}>0$ and $\uv_m^{k+1}\vv_i^{k+1}<0$ (flipping these inequalities yields a symmetric situation where the feature will first decrease and then increase). The duration of the illusory correlation can be estimated by contrasting the time at which the first mode is learned compared to the second. In particular, suppose the second mode's singular value is smaller by an amount $\Delta$, that is, $s_{k+1}=s_k-\Delta$. Then the illusory correlation persists for a time
\begin{eqnarray*}
	t_{k+1}-t_k & = &  \frac{\tau}{s_k-\Delta}\ln \frac{s_k-\Delta}{\epsilon} - \frac{\tau}{s_k}\ln \frac{s_k}{\epsilon}\\
		 	 %& = & \frac{\tau s_k\ln \frac{s_k-\Delta}{\epsilon} - (s_k-\Delta)\ln \frac{s_k}{\epsilon}}{s_k(s_k-\Delta)}\\
			 & = & \frac{\tau s_k\ln \frac{s_k-\Delta}{s_k} +\tau \Delta\ln \frac{s_k}{\epsilon}}{s_k(s_k-\Delta)}\\
			 & \approx & \frac{ \tau \Delta\ln \frac{s_k}{\epsilon}}{s_k^2} 
\end{eqnarray*}
in the regime where $\epsilon \ll \Delta \ll s_k$, or approximately a period of length $O(\Delta)$. While illusory correlations can cause the error on one specific feature to increase, we note that the total error across all features and items always decreases or remains constant (as is the case for any gradient descent procedure).

In contrast, the shallow network exhibits no illusory correlations. The prediction for feature $m$ on item $i$ is 
\begin{eqnarray*}
\hat \y_m^i & = &\sum_\alpha b_\alpha(t)\uv_m^\alpha \vv_i^\alpha \\
& = & \sum_\alpha \left[ s_\alpha\left(1-e^{-t/ \tau}\right)   + b_\alpha^0e^{-t/\tau}\right]\uv_m^\alpha \vv_i^\alpha \\
& = & \left(1-e^{-t/ \tau}\right) \underbrace{\left[\sum_\alpha s_\alpha\uv_m^\alpha \vv_i^\alpha\right]}_{c_1} + e^{-t/\tau}\underbrace{\sum_\alpha b_\alpha\uv_m^\alpha \vv_i^\alpha}_{c_2}\nonumber \\
%& = & c_1\left(1-e^{-t/ \tau}\right) + c_2e^{-t/\tau} \\
& = & c_1-(c_1-c_2)e^{-t/ \tau}
\end{eqnarray*}
which is clearly monotonic in $t$. Therefore shallow networks never yield illusory correlations where the sign of the progress on a particular feature changes over the course of learning. 

\section{Organizing and Encoding Knowledge}

\subsection{Category membership, typicality, prototypes}
The singular value decomposition satisfies a set of mutual constraints that provide consistent relationships between category membership and item and feature typicality. In particular, form the matrix $\bo=[\y^1 \cdots \y^{N_1}]$ consisting of the features of each object in its columns. We assume that the input here directly codes for object identity using a one-hot input vector ($X=I$). Then the input-output correlation matrix which drives learning is $\sio=E[\y\x^T]=\frac{1}{P}\bo$. The dynamics of learning are thus driven by the singular value decomposition of $\bo$,
\begin{equation}
	\frac{1}{P}\bo  =  \ls \sv \rs^T, \label{typicality_svd}
\end{equation}
where the matrices of left and right singular vectors are orthogonal ($\ls^T\ls=I$ and $\rs^T\rs=I$). Because of this orthogonality, multiplying both sides by $\sv^{-1}\ls^T$ from the left we have,
\begin{eqnarray*}
	\frac{1}{P}\sv^{-1}\ls^T\bo  & = & \sv^{-1}\ls^T\ls\sv\rs^T,\\
	\frac{1}{P}\sv^{-1}\ls^T\bo  & = & \rs^T
\end{eqnarray*}
Pulling out the element at the $i$th row and the $\alpha$th column of $\rs$ on both sides, we obtain Eqn.\typicality of the main text,
\begin{equation*}
	\vv^\alpha_i =  \frac{1}{Ps_\alpha} \sum_{m=1}^{N_3}  \uv^\alpha_m \ov^{i}_m.
\end{equation*}

Similarly, multiplying Eqn.\,\req{typicality_svd} from the right by $\rs\sv^{-1}$ yields,
\begin{eqnarray*}
	\frac{1}{P}\bo \rs\sv^{-1}  & = & \ls \sv \rs^T\rs\sv^{-1},\\
	\frac{1}{P}\bo \ls\sv^{-1} & = & \ls.
\end{eqnarray*}
Extracting the elements at the $i$th row and $\alpha$th column yields Eqn.\prototype of the main text,
\begin{equation}
\uv^\alpha_m =   \frac{1}{Ps_\alpha}\sum_{i=1}^{N_1}  \vv^\alpha_i \ov^{i}_m. \nonumber
\end{equation}
\subsection{Category coherence}

Real world categories may be composed of a small number of items and features amid a large background of many items and possible features which do not possess category structure. Here we consider the task of identifying disjoint categories in the presence of such noise. We show that a single category coherence quantity determines the speed and accuracy of category recovery by a deep linear neural network, and compute the threshold category coherence at which deep linear networks begin to correctly recover category structure.

We consider a dataset of $N_o$ objects and $N_f$ features, in which a category of $K_o$ objects and $K_f$ features is embedded. That is, a subset $C_f$ of $K_f=|C_f|$ features occur with high probability $p$ for the subset $C_i$ of $K_o=|C_i|$ items in the category. Background features (for which either the feature or item are not part of the category) occur with a lower probability $q$. Define the random matrix $\br$ of size $N_f \times N_o$ to have entries $\br_{ij}=1$ with probability $p$ and $0$ with probability $1-p$ provided $i\in C_f$ and $j \in C_i$, and $ \br_{ij}=1$ with probability $q$ and $0$ with probability $1-q$ otherwise. A realization of this matrix yields one environment containing items and features with a category embedded into it. To access general properties of this setting, we study the behavior in the high-dimensional limit where the number of features and items is large, $N_f,N_o\rightarrow \infty$, but their ratio is constant, $N_o/N_f\rightarrow c \in (0,1]$.

We suppose that the features are recentered and rescaled such that background features have zero mean and variance $1/N_f$ before being passed to the network. That is, we define the normalized, rescaled feature matrix
\begin{equation}
    \tilde \br = \frac{1}{\sqrt{N_f q(1-q)}}(\br - q{\bf 1 1^T})
\end{equation}
where we have used the fact that $E[y_i]=q$ and $Var[y_i]=q(1-q)$ for a background feature $i$  to derive the appropriate rescaling. With this rescaling we can approximately rewrite $\tilde \br$ as a random matrix perturbed by a low rank matrix corresponding to the embedded category,
\begin{equation}
    \tilde \br \approx \bx + {\bf P}.
\end{equation}
Here each element of the noise matrix $\bx$ is independent and identically distributed as $\bx_{ij}=\frac{1}{\sqrt{N_f q(1-q)}}(x - q)$ where $x$ is a Bernoulli random variable with probability $q$. The signal matrix ${\bf P}$ containing category information is low rank and given by
\begin{equation}
    {\bf P} = \theta\frac{1}{\sqrt{K_fK_o}} {\bf 1}_{C_f}{\bf 1}_{C_i}^T \label{eq:perturb_mtx}
\end{equation}
where ${\bf 1}_{C}$ is a vector with ones on indices in the set $C$ and zeros everywhere else, and $\theta$ is the associated singular value of the low rank category structure. In particular, elements of $\br$ for items and features within the category have a mean value of $p$. Hence using this and applying the mean shift and rescaling as before, we have
\begin{eqnarray}
    \theta & = & \frac{(p-q)\sqrt{K_fK_o}}{\sqrt{N_f q(1-q)}}.
\end{eqnarray}

To understand learning dynamics in this setting, we must compute the typical singular values and vectors of $\tilde \br$. Theorem 2.9 of \cite{Benaych-georges2012} states that recovery of the correct singular vector structure only occurs for signal strengths above a threshold (an instance of the BBP phase transition \cite{Baik2005}). In particular, let $\tilde \uv, \tilde \vv$ be the feature and object analyzer vectors of $\tilde \br$ respectively (left and right singular vectors respectively), and let
\begin{eqnarray}
    \uv^{\textrm{Ideal}}& =& \frac{1}{\sqrt{K_f}}{\bf 1}_{C_f},\\
    \vv^{\textrm{Ideal}} &= &\frac{1}{\sqrt{K_o}}{\bf 1}_{C_o}
\end{eqnarray} 
be the ground truth feature and object analyzers arising from the category structure in Eqn.~\req{eq:perturb_mtx}. Then 
\begin{eqnarray}
    \left(\tilde \uv^T\uv^{\textrm{Ideal}}\right)^2&\xrightarrow{a.s.}&
    \begin{cases}
      1 - \frac{c + \theta^2}{\theta^2(\theta^2 + 1)} & \text{for } \theta > c^{1/4}\\    
      0 & \text{otherwise}   
\end{cases} \label{eq:feat_analy_overlap}\\
\left(\tilde \vv^T\vv^{\textrm{Ideal}}\right)^2&\xrightarrow{a.s.}&
    \begin{cases}
      1 - \frac{c(1+\theta^2)}{\theta^2(\theta^2 + c)} & \text{for } \theta > c^{1/4}\\    
      0 & \text{otherwise}   
\end{cases} \label{eq:obj_analy_overlap}
\end{eqnarray}
where $a.s.$ denotes almost sure convergence (i.e., with probability 1) in the high-dimensional limit ($N_f,N_o \rightarrow \infty$ and $N_o/N_f=c$).

In essence, for $\theta \leq c^{1/4}$, the learned feature and object analyzer vectors will have no overlap with the correct category structure. For $\theta > c^{1/4}$, the feature and object analyzer vectors will have positive dot product with the true category structure yielding at least partial recovery of the category. Using the definitions of $\theta$ and $c$ and straightforward algebra, the recovery condition $\theta > c^{1/4}$ can be written as
\begin{equation}
    \frac{(p-q)^2K_fK_o}{ q(1-q)\sqrt{N_fN_o}} > 1.
\end{equation}
This motivates defining category coherence as
\begin{eqnarray}
    \mathcal{C}& \equiv & \frac{(p-q)^2K_fK_o}{ q(1-q)\sqrt{N_fN_o}}\\
     & = & \textrm{SNR}\frac{K_fK_o}{\sqrt{N_fN_o}}
\end{eqnarray}
where we have defined the signal-to-noise ratio $\textrm{SNR} = \frac{(p-q)^2}{q(1-q)}$. 

So defined, for a fixed item/feature ratio $c$, the category coherence $\mathcal{C}$ completely determines the performance of category recovery. To see this, we note that $\theta^2=c^{1/2}\mathcal{C}$ such that
%\begin{eqnarray}
%    \theta^2& = & \frac{(p-q)^2K_fK_o}{N_f q(1-q)} \\
%    & = & \sqrt{\frac{N_o}{N_f}}\frac{(p-q)^2K_fK_o}{\sqrt{N_fN_o} q(1-q)} \\
%    & = & \sqrt{c}\mathcal{C}
%\end{eqnarray}. 
Eqns. \req{eq:feat_analy_overlap}-\req{eq:obj_analy_overlap} can be written as
\begin{eqnarray}
    \left(\tilde \uv^T\uv^{\textrm{Ideal}}\right)^2&\xrightarrow{a.s.}&
    \begin{cases}
      1 - \frac{1 + c^{-1/2}\mathcal{C}}{\mathcal{C}(\mathcal{C} + c^{-1/2})} & \text{for } \mathcal{C} > 1\\    
      0 & \text{otherwise}   
\end{cases} \label{eq:feat_analy_overlap_catcoh}\\
\left(\tilde \vv^T\vv^{\textrm{Ideal}}\right)^2&\xrightarrow{a.s.}&
    \begin{cases}
      1 - \frac{1+c^{1/2}\mathcal{C}}{\mathcal{C}(\mathcal{C} + c^{1/2})} & \text{for } \mathcal{C} > 1\\    
      0 & \text{otherwise}   
\end{cases} \label{eq:obj_analy_overlap_catcoh}
\end{eqnarray}
Hence recovery of category structure can be described by a single category coherence quantity that is sensitive to both the signal-to-noise ratio of individual features in the category relative to background feature variability, weighted by the size of the category. Finally, we reiterate the regime of validity for the analysis presented here: the theory applies in the limit where $N_f$ and $N_o$ are large, the ratio $c=N_o/N_f\in(0,1]$ is finite (implying $N_f > N_o$), and the category size is on the order of the square root of the total number of items and features, $K_fK_o  \sim  \sqrt{N_fN_o}$.

\subsection{Basic categories}

To generalize the notion of category coherence further, we propose to simply define category coherence as the singular value associated with a categorical distinction in the SVD of the input-output correlations $\sio$. In this section we show that this definition can give rise to a basic level advantage depending on the similarity structure of the categories, and gives rise to an intuitive notion of category coherence based on within-category similarity and between-category difference. We additionally show that this definition makes category coherence dependent on the global structure of the dataset, through a well-known optimality condition. 

\textit{Hierarchical singular values from item similarities.} The hierarchical generative model considered previously has a simple structure of independent diffusion down the hierarchy. This results in singular values that are always a decreasing function of the hierarchy level. Here we show how more complex (but still hierarchical) similarity structures between items can give rise to a basic level advantage; and that defining category coherence as the associated singular value for a categorical distinction recovers intuitive notions of category coherence.

Suppose we have a set of items with input-output correlation matrix $\sio$. The singular values are the square root of the eigenvalues of the item similarity matrix,
\begin{equation}
    {\sio}^T\sio\equiv \so,
\end{equation}
and the object analyzer vectors $\vv^\alpha,~ \alpha=1,\cdots,P$ are the eigenvectors of $\so$. We assume that the object analyzer vectors exactly mirror the hierarchical structure of the items, and for simplicity focus on the case of a regularly branching tree.

By assumption, the item similarity matrix has decomposition
\begin{equation}
    \so = \sum_{\alpha=1}^P \lambda_\alpha \vv^\alpha {\vv^\alpha}^T.
\end{equation}
As described previously, eigenvectors come in groups corresponding to each hierarchical level $k$.  

In this setting, the similarity matrix will have a hierarchical block structure (as can be seen in \figbasiclvl~of the main text). Each block corresponds to a subset of items, and blocks are either disjoint (containing different items) or nested (one block containing a subset of the items of the other). The blocks are in one to one correspondence with a rooted regularly branching tree, with leaves corresponding to each item and one block per internal node. Each block corresponding to a node of the tree at level $k$ has constant entries of
\begin{equation}
	q_k=\frac{1}{N_3}\sum^{N_3}_{i=1}\y_i^{\mu_1}\y_i^{\mu_2},
\end{equation}
the similarity between any two items $\mu_1,\mu_2$ with closest common ancestor at level $k$.

The eigenvalue associated with a category $C$ at level $k$ in the hierarchy can be written as
\begin{equation}
    \lambda_k = \sum_{j \in C} \so_{ij} - \sum_{j\in S(C)}\so_{ij}  \quad \textrm{ for any } i \in C \label{tree_cat_coh}
\end{equation}
where $S(C)$ is any sibling category of $C$ in the tree (i.e.~another category at the same hierarchical level).
That is, take any member $i$ of category $C$, and compute the sum of its similarity to all members of category $C$ (including itself); then subtract the similarity between member $i$ and all members of one sibling category $S(C)$. Hence this may directly be interpreted as the total within category similarity minus the between category difference. 

A basic level advantage can thus occur if between category similarity is negative, such that items in different categories have anticorrelated features. This will cause the second term of Eqn.~\req{tree_cat_coh} to be positive, boosting category coherence at that level of the hierarchy. The category coherence of superordinate categories will decrease (because within category similarity will decrease), and subordinate categories will be unaffected. If the anticorrelation is strong enough, an intermediate level can have higher category coherence, and be learned faster, than a superordinate level.

\textit{Global optimality properties of the SVD.} Our proposal to \textit{define} the category coherence $\mathcal{C}$ as the associated singular value for a particular object-analyzer vector makes category coherence fundamentally dependent on the interrelations between all items and their properties. To see this, we observe that the singular value decomposition obeys a well-known \textit{global} optimality condition: if we restrict our representation of the environment to just $k$ linear relations, then the first $k$ modes of the SVD yield the lowest total prediction error of all linear predictors. In particular, suppose the network retains only the top $k$ modes of the singular value decomposition, as would occur if training is terminated early before all modes have risen to their asymptote. The network predicts features $\tilde \bo = \ls \tilde \sv \rs^T$, where $\tilde \sv$ contains just the first $k$ singular values with the remaining diagonal elements set to zero (that is, $\tilde \sv_{ii} = \sv_{ii}$ for $i\leq k$ and $\tilde \sv_{ii} = 0$ otherwise). The Eckart--Young--Mirsky theorem states that $\tilde \bo$ is a solution to
\begin{equation}
	\min_{\bb,\textrm{rank}(\bb)\leq k} \left\| \bo - \bb \right\|_F.
\end{equation}
Hence out of all rank $k$ representations of the environment, the truncated SVD yields the minimum total error. 

In the terminology of deep linear neural networks, out of all networks with $N_2= k$ or fewer hidden neurons, networks with total weights $\tilde \wb\tilde \wa=\tilde\bo=\ls\tilde \sv \rs^T$ are minimizers of the total sum squared error,
\begin{equation*}
	\min_{\wa,\wb,N_2\leq k} SSE(\wa,\wb).
\end{equation*}
We include a proof of this fact for completeness. First, note that 
\begin{eqnarray}
	SSE(\tilde \wa, \tilde \wb) = \frac{1}{2}||\ls(\sv - \tilde \sv)\rs||_F^2 = \frac{1}{2}\sum_{i=k+1}^{N_1} s_i(\bo)^2. \label{sse_frobenius}
\end{eqnarray}
where here and in the following we will denote the $i$th largest singular value of the matrix $A$ as $s_i(A)$. For two matrices $\bc\in R^{N_3 \times N_1}$ and $\bd\in R^{N_3 \times N_1}$,  Weyl's theorem for singular values states that
\begin{equation*}
	s_{i+j-1}(\bc+\bd) \leq s_i(\bc) + s_j(\bd)
\end{equation*}
for $1\leq i,j \leq N_1$ and $i+j-1\leq N_1$. Taking $j=k+1$, $\bc=\bo-\wb\wa$, and $\bd=\wb\wa$ yields
\begin{eqnarray}
	s_{i+k}(\bo)& \leq& s_i(\bo-\wb\wa) + s_{k+1}(\wb\wa)  \label{sv_weyl_a}\\
	& \leq &  s_i(\bo-\wb\wa) \label{sv_weyl_b}
\end{eqnarray}
for $1 \leq i \leq N_1-k$. In the last step we have used the fact that $s_{k+1}(\wb\wa)=0$ since $\wb\wa$ has rank at most $k$. We therefore have
\begin{eqnarray}
	\frac{1}{2}||\bo - \wb\wa||_F^2 &=& \frac{1}{2}\sum_{i=1}^{N_1}s_i(\bo-\wb\wa)^2 \nonumber\\
	&\geq& \frac{1}{2}\sum_{i=1}^{N_1-k}s_i(\bo-\wb\wa)^2 \label{sv_eckart_a}\\
	& \geq & \frac{1}{2}\sum_{i=k+1}^{N_1}s_i(\bo)^2 \label{sv_eckart_b}\\
	&=&\frac{1}{2} ||\bo - \tilde \wb \tilde\wa||_F^2, \nonumber
\end{eqnarray}
where from \req{sv_eckart_a}-\req{sv_eckart_b} we have used \req{sv_weyl_a}-\req{sv_weyl_b} and the last equality follows from Eqn.~\req{sse_frobenius}.
Hence 
\begin{eqnarray}
	SSE(\tilde \wa, \tilde \wb) \leq SSE(\wa,\wb)
\end{eqnarray}
as required.

As a simple example of how local changes to the features of a few items can cause global reorganization of categorical structure in the SVD, we consider the hierarchical dataset from \figsvd \,in the main text, but add a single additional feature. If this feature is not possessed by any of the items, then the categorical decomposition reflects the hierarchical structure of the dataset as usual. However if this feature is possessed by both the \textit{Canary} and \textit{Rose} (perhaps a feature like \textit{Brightly colored}), the resulting categorical structure changes substantially, as shown in \sfigglobalopt. While the highest two levels of the hierarchy remain similar, the lowest two levels have been reconfigured to group the \textit{Canary} and \textit{Rose} in one category and the \textit{Salmon} and \textit{Oak} in another. Consider, for instance, the \textit{Salmon}: even though its own feature vector has not changed, its assignment to categories has. In the original hierarchy, it was assigned to a \textit{bird-fish} distinction, and did not participate in a \textit{tree-flower} distinction. With the additional feature, it now participates in both a \textit{bright-dull} distinction and another distinction encoding the differences between the \textit{Canary/Oak} and \textit{Salmon/Rose}. Hence the mapping between features and categorical structure implied by the SVD can be non-local and nonlinear: small perturbations of the features of items can sometimes result in large changes to the singular vectors. This specific example is not intended to be an actual description of the property correlations for these items. Rather, we use it narrowly to demonstrate the point that the categorical structure arising from the SVD is a global property of all items and their features, and the categorical structure applied to one specific item can be altered by the features of other items.

\begin{figure}
\begin{center}
\includegraphics[width=2.4in]{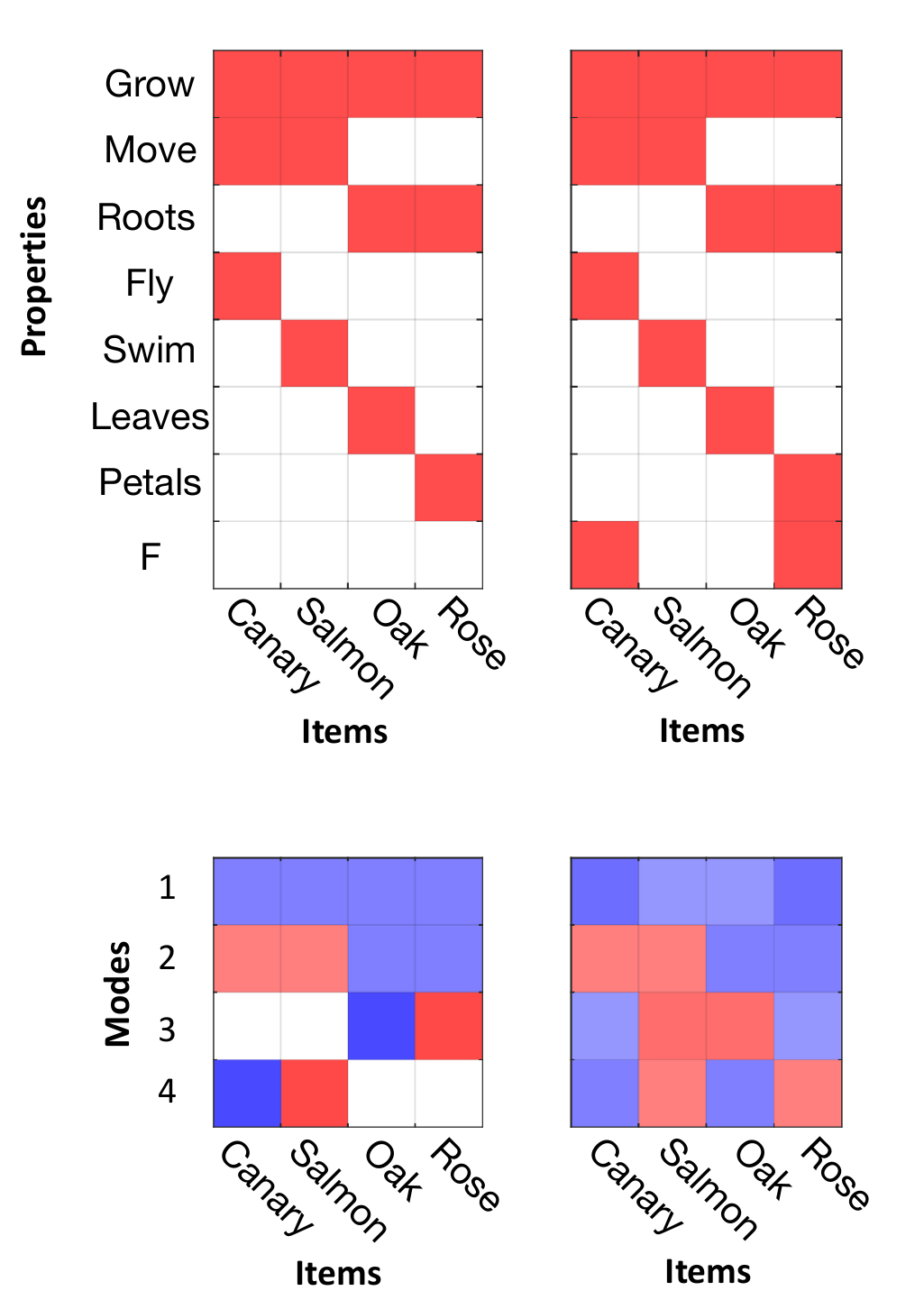}
\end{center}

\caption{Category structure is a nonlocal and nonlinear function of the features. Left column: a toy dataset with hierarchical structure (top) has object analyzer vectors that mirror the hierarchy (bottom). Right column: Adding a new feature \textit{F} to the dataset (top) causes a substantial change to the category structure (bottom). In particular the features of the \textit{Salmon} are identical in both datasets, yet the categorical groupings the \textit{Salmon} participates in have changed, reflecting the fact that the SVD is sensitive to the global structure of the dataset. }
\end{figure}

\subsection{Discovering and representing explicit structures}

To investigate how datasets with certain underlying structural forms come to be represented in the neural network, we consider drawing datasets from probabilistic graphical models specified by graphs over items (and possibly hidden variables). To go from a graph to feature values for each item, we follow \cite{Kemp2008} and use a Gaussian Markov random field. Intuitively, this construction causes items which are nearby in the graph to have more similar features. 

In particular, consider a graph consisting of a set of nodes $\mathcal V$ of size $K=|\mathcal V|$, connected by a set of undirected edges $\mathcal E$ with lengths $\{e_{ij}\}$, where $e_{ij}$ is the length of the edge between node $i$ and node $j$. Each item in the environment is associated with one node in the graph, but there can be more nodes than items. For instance, a tree structure has nodes for each branching point of the tree, but items are associated only with the leaves (in \figstructform~of the main text, nodes associated with items are depicted as filled circles, while unassociated nodes lie at edge intersections). We construct the $K \times K$ weighted adjacency matrix ${\bf A}$ where ${\bf A}_{ij}=1/e_{ij}$ and ${\bf A}_{ij}=0$ if there is no edge between nodes $i$ and $j$. Next, we form the graph Laplacian ${\bf L}= {\bf D}-{\bf A}$ where ${\bf D}$ is the diagonal weighted degree matrix with ${\bf D}_{ii}=\sum_{j=1}^K {\bf A}_{ij}$. We take the value of a particular feature $m$ across nodes in the graph to be distributed as
\begin{equation*}
	\tilde {\bf f} \sim \mathcal{N}\left(0, \tilde {\bf \Phi}^{-1}\right) \label{gmrf_feat}
\end{equation*}
where $\tilde {\bf f}$ is a length $K$ vector of feature values for each node, and $\tilde {\bf \Phi}={\bf L} + 1/\sigma^2 {\bf I}$ is the precision matrix (inverse covariance matrix) of the Gaussian distribution. Here the parameter $\sigma^2$ instantiates graph-independent variation in the feature values which ensures the inverse exists. Finally, to obtain a length $P$ vector {\bf f} of feature values across items (rather than across all nodes in the graph) we take the subset of the vector $\tilde {\bf f}$ corresponding to nodes with associated items. This can be written as ${\bf f} = {\bf M}\tilde {\bf f}$ for an appropriate matrix ${\bf M}\in \mathbb{R}^{P \times K}$ which has ${\bf M}_{ij}=1$ if item $i$ is associated with node $j$ and is zero otherwise. This is a linear transformation of a Gaussian, and hence ${\bf f}$ is Gaussian zero mean with covariance $ {\bf \Phi}^{-1}={\bf M}\tilde {\bf \Phi}^{-1}{\bf M}^T$, 
\begin{equation}
	 {\bf f} \sim \mathcal{N}\left(0,  {\bf \Phi}^{-1}\right). \label{gmrf_feat2}
\end{equation}To obtain multiple features, we assume that features are drawn independently according to Eq. \req{gmrf_feat2}. 

This approach describes a generation process for a dataset: A set of $N_3$ features are drawn, yielding the dataset $\{\x^i,\y^i\}, ~i=1,\ldots,P$ where for simplicity we assign one-hot input vectors $\x^i$ to each item such that ${\bf X} = [\x^1 \cdots \x^P]={\bf I}$. This dataset is then presented to the neural network for training, and the dynamics are driven by the SVD of $\sio=\frac{1}{P}\sum_i^P \y^i\x^{i^T}=\frac{1}{P}\bf{YX}^T$ where ${\bf Y}=[\y^1 \cdots \y^P]$ is the concatenated matrix of features. From the definition of the SVD, the object analyzer vectors are the eigenvectors of the matrix 
\begin{eqnarray*}
	{\sio}^T\sio& =& \frac{1}{P^2}{\bf XY}^T{\bf YX}\\
	& = & \frac{1}{P^2}{\bf Y}^T{\bf Y} \equiv  \so.\\
\end{eqnarray*}
Now we note that 
\begin{eqnarray*}
\so_{ij} & = & \frac{1}{P^2} \y^{i^T} \y^{j} \\
 & = & \frac{1}{P^2} \sum_{m=1}^{N_3} \y^{i}_m \y^j_m \\
 & = & \frac{N_3}{P^2} \left( \frac{1}{N_3}\sum_{m=1}^{N_3} \y^{i}_m \y^j_m \right).
\end{eqnarray*}
As the number of features grows $(N_3\rightarrow \infty)$, this sample average converges to 
\begin{eqnarray*}
 \so_{ij} & = & \frac{N_3}{P^2} E\left[ {\bf f}^{i} {\bf f}^j \right] 
 \end{eqnarray*}
 and hence from Eq.~\req{gmrf_feat2}, 
 \begin{eqnarray*}
 \so=\frac{N_3}{P^2} {\bf \Phi}^{-1}.
 \end{eqnarray*}

Up to a scalar, the item covariance matrix is simply the covariance structure arising from the graph; and because matrix inversion preserves eigenvectors, the eigenvectors of the matrix ${\bf \Phi}$ are the object analyzer vectors. Finally, the singular values are $s_\alpha = \frac{\sqrt{N_3}}{P\sqrt{\zeta_\alpha}}$, where $\zeta_\alpha$ is the $\alpha$'th eigenvalue of ${\bf \Phi}$.

We now describe how the specific graph types considered in the main text result in structured matrices for which the SVD may be calculated analytically.
\begin{description}
	\item[Clusters] Here we consider partitioning $P$ items into a set of $N_c\leq P$ clusters, yielding a graph in which each item in a cluster $b$ is connected by a constant length edge $e_b$ to a hidden cluster identity node. Let $M_b$ be the number of items in cluster $b$. It is easy to see that the resulting item correlation structure is block diagonal with one block per cluster; and each block has the form ${\bf \Phi}^{-1}_b=c_1^b{\bf 1}+c_2^b{\bf I}$ where ${\bf 1} \in \mathbb{R}^{M_b \times M_b}$ is a constant matrix of ones, ${\bf I}$ is an identity matrix, $b=1,\cdots,N_c$ is the block index, and $c^b_1,c^b_2$ are scalar constants

	\begin{eqnarray*}
	    c^b_1 & = & \frac{\sigma^2}{M_b+1} + \frac{M_b-1}{(1/e_b + 1/\sigma^2)M_b}\\
	    &&+ \frac{1}{((M_b+1)/e_b+1/\sigma^2)M_b(M_b+1)}\\
	    c^b_2 & = & \frac{\sigma^2}{M_b+1} - \frac{1}{M_b(1/e_b+1/\sigma^2)} \\
	    & & + \frac{1}{((M_b+1)/e_b + 1/\sigma^2)M_b(M_b+1)}
	\end{eqnarray*}
	To understand learning dynamics in this setting, we must compute the eigenvalues and eigenvectors of this correlation structure. The eigenvalues and eigenvectors of a block diagonal matrix are simply the concatenated eigenvalues and eigenvectors of each of the blocks (where the eigenvectors from a block are padded with zeros outside of that block). Looking at one block $b$, the constant vector ${\bf v}=1/\sqrt{M_b}{\bf 1}$ is an object analyzer vector with eigenvalue 
	\[s_1=\frac{1+M_b\sigma^2/e_b}{(M_b+1)/e_b+1/\sigma^2}.\] 
	The remaining $M_b-1$ eigenvalues are all equal to
	\[s_2=\frac{1}{1/e_b+1/\sigma^2}.\]

	 From these results we can draw several conclusions about the speed of learning simple category structure. First, we note that the shared structure in a category, encoded by the constant eigenvector, is always more prominent (and hence will be learned faster) than the item-specific information. That is, $s_1$ is always larger than $s_2$ in the relevant regime $M_b\geq2$, $e_b>0$, and $\sigma>0$. To see this, we differentiate the difference $s_1-s_2$ with respect to $M_b$ and $e_b$ and set the result to zero to find extremal points. This yields $M_b=0$, and $1/e_b=0$ or $1/e_b=-2/(M_b\sigma^2 + 2\sigma^2)$. Hence there  are no critical points in the relevant region, and we therefore test the boundary of the constraints. For $e_b\rightarrow 0$, we have $s_1-s_2=\frac{\sigma^2}{1 + 1/M_b}$ which is increasing in $M_b$. For $M_b=2$, we have $s_1-s_2=\frac{2\sigma^6}{3\sigma^4+4\sigma^2e_b + e_b^2}$ which is decreasing in $e_b$. The minimum along the boundary would thus occur at $M_b=2, e_b\rightarrow\infty$, where the difference converges to zero but is positive at any finite value. Testing a point in the interior yields a higher value (for instance $M_b=3$ and $e_b=1$ yields $s_1-s_2=\frac{3\sigma^6}{4\sigma^4 + 5\sigma^2+1}\geq 0$), confirming that this is the global minimum and $s_1 > s_2$ in this domain. Hence categorical structure will typically be learned faster than idiosyncratic item-specific information.

	We note that the graph we have constructed is only one way of creating categorical structure, which leaves different clusters independent. In particular, it establishes a scenario in which features of members in each category are positively correlated, but features of members of different categories are simply not correlated, rather than being anticorrelated. Hence the model considered instantiates within-cluster similarity, but does not establish strong between-cluster difference. We note that such anticorrelations can be readily incorporated by including negative links between hidden cluster nodes. 
	
	For the results presented in \figstructform~we used $N_c=3$ clusters with $M_b=\{4,2,3\}$ items per cluster, $e_b=0.24$ for all clusters, and $\sigma=4$.
	
	\item[Trees]
	To construct a dataset with an underlying tree structure, in our simulations we make use of the hierarchical branching diffusion process described previously. Specifically, we used a three level tree with binary branching and flip probability $\epsilon=.15$. As shown, this gives rise to a hierarchically structured singular value decomposition. 
	
	To understand the generality of this result we can also formulate hierarchical structure in the Gaussian Markov random field framework. To implement a tree structure, we have a set of internal nodes corresponding to each branching point in the tree, in addition to the $P$ leaf nodes corresponding to individual items. We form the adjacency graph ${\bf A}$ and compute the inverse precision matrix ${\tilde \Phi}$ as usual. To obtain the feature correlations on just the items of interest, we project out the internal nodes using the linear map ${\bf M}$. This ultimately imparts ultrametric structure in the feature correlation matrix $\so$. As shown in \cite{Khrennikov2005}, such matrices are diagonalized by the ultrametric wavelet transform, which therefore respects the underlying tree structure in the dataset. An important special case is binary branching trees, which are diagonalized by the Haar wavelets \cite{Murtagh2007}.
	
	\item[Rings and Grids]
	
	Items arrayed in rings and grids, such as cities on the globe or locations in an environment, yield correlation matrices with substantial structure. For a ring, correlation matrices are circulant, meaning that every row is a circular permutation of the preceding row. For a grid, correlation matrices are Toeplitz, meaning that they have constant values along each diagonal. Circulant matrices are diagonalized by the unitary Fourier transform \cite{Gray2005}, and so object analyzer vectors will be sinusoids of differing frequency. The associated singular value is the magnitude of the Fourier coefficient. If correlations are decreasing with distance in the ring, then the broadest spatial distinctions will be learned first, followed by progressive elaboration at ever finer scales, in an analogous process to progressive differentiation in hierarchical structure. Grid structures are not exactly diagonalized by the Fourier modes, but the eigenvalues of Circulant and Toeplitz matrices converge as the grid structure grows large and edge effects become small \cite{Gray2005}. Our example is given in a 1D ring, but the same structure arises for higher dimensional structure (yielding, for instance, doubly block circulant structure in a 2D ring \cite{Gray2005,Tee2005} which is diagonalized by the 2D Fourier transform).
	
	In \figstructform~, we used $P=20$ items in a ring-structured GMRF in which items are only connected to their immediate neighbors. These connections have length $e_{ij}=1/.7$ such that ${\bf A}_{ij}=0.7$ if $i,j$ are adjacent nodes. Finally we took the individual variance to be $1/\sigma^2=0.09$.
	
	\item[Orderings]
	    A simple version of data with an underlying transitive ordering is given by a 1D chain. In the GMRF framework, this will yield Toeplitz correlations in which the first dimension encodes roughly linear position as described above for grids. To instantiate a more complex example, in \figstructform~we also consider a domain in which a transitive ordering is obeyed exactly: any feature possessed by a higher order entity is also possessed by all lower-order entities. This situation might arise in social dominance hierarchies, for example, with features corresponding to statements like ``individual $i$ dominates individual $j$'' (see for example \cite{Kemp2008,Kemp2009}). To instantiate this, we use the input-output correlations
	    	\begin{equation}
	    \sio=\frac{1}{P}\begin{bmatrix}
1 & 0 & 0 & 0 & 0 & 0 & 0 & 0 & 0\\ 
1 & 1 & 0 & 0 & 0 & 0 & 0 & 0 & 0\\ 
1 & 1 & 1 & 0 & 0 & 0 & 0 & 0 & 0\\ 
1 & 1 & 1 & 1 & 0 & 0 & 0 & 0 & 0\\ 
1 & 1 & 1 & 1 & 1 & 0 & 0 & 0 & 0\\ 
1 & 1 & 1 & 1 & 1 & 1 & 0 & 0 & 0\\ 
1 & 1 & 1 & 1 & 1 & 1 & 1 & 0 & 0\\ 
1 & 1 & 1 & 1 & 1 & 1 & 1 & 1 & 0\\ 
1 & 1 & 1 & 1 & 1 & 1 & 1 & 1 & 1\\ 
	    \end{bmatrix},
	\end{equation}
	which realizes a scenario in which a group of items obeys a perfect transitive ordering. This structure yields feature correlations that take constant values on or below the diagonal in each column, and on or to the right of the diagonal in each row. 
	
	\item[Cross-cutting Structure]
	Real world datasets need not conform exactly to any one of the individual structures described previously. The domain of animals, for instance, might be characterized by a broadly tree-like structure, but nevertheless contains other regularities such as \textit{male/female}, \textit{predator/prey}, or \textit{arctic/equatorial} which can cut across the hierarchy \cite{Mcclelland2016}.  These will be incorporated into the hidden representation as additional dimensions which can span items in different branches of the tree. The example given in \figstructform~instantiates a version of this scenario. The input-output correlation matrix is given by 
	\begin{equation}
	    \sio=\frac{1}{P}\begin{bmatrix}
	        1 & 1 & 1 & 1 & 1 & 1 & 1 & 1\\ 
            1 & 1 & 1 & 1 & 0 & 0 & 0 & 0\\ 
            0 & 0 & 0 & 0 & 1 & 1 & 1 & 1\\ 
            1.1 & 1.1 & 0 & 0 & 0 & 0 & 0 & 0\\ 
            0 & 0 & 1.1 & 1.1 & 0 & 0 & 0 & 0\\ 
            0 & 0 & 0 & 0 & 1.1 & 1.1 & 0 & 0\\ 
            0 & 0 & 0 & 0 & 0 & 0 & 1.1 & 1.1\\ 
            1.1 & 1.1 & 0 & 0 & 0 & 0 & 0 & 0\\ 
            0 & 0 & 1.1 & 1.1 & 0 & 0 & 0 & 0\\ 
            0 & 0 & 0 & 0 & 1.1 & 1.1 & 0 & 0\\ 
            0 & 0 & 0 & 0 & 0 & 0 & 1.1 & 1.1\\ 
            1 & 0 & 1 & 0 & 1 & 0 & 1 & 0\\ 
            0 & 1 & 0 & 1 & 0 & 1 & 0 & 1
	    \end{bmatrix}.
	\end{equation}
	This dataset has a hierarchical structure that is repeated for pairs of items, except for the final two features which encode categories that cut across the hierarchy. The feature values of $1.1$ which occur in the finer levels of the hierarchy are to create a separation in singular values between the hierarchy modes and the cross-cutting structure.
	
	Examples of this form can be cast in the GMRF framework by combining a tree structure with links to two categories representing the cross-cutting dimensions. The structure of the graph is depicted approximately in ~\figstructform, but we note that it cannot be accurately portrayed in three dimensions: all members of each cross-cutting category should connect to a latent category node, and the length of the links to each category member should be equal. Additionally, the final links from the tree structure (depicted with dashed lines) should have length zero, indicating that without the cross-cutting structure the paired items would not differ. 
	
\end{description}

\section{Deploying Knowledge: Inductive Projection}
In this section we consider how knowledge about novel items or novel properties will be extended to other items and properties. For instance, suppose that a novel property is observed for a familiar item (e.g., ``a \textit{pine} has property \textit{x}''). How will this knowledge be extended to other items (e.g., ``does a \textit{rose} have property \textit{x}'')? Here we are interested in the interaction between two timescales of learning: the slow, gradual process of development that we describe with error correcting learning in a deep linear network; and the potentially rapid generalizations that can be made upon learning a new fact, based on the current background knowledge embedded in the network. To model this more rapid learning of a new fact, we add a new neuron representing the novel property or item, and following \cite{Rogers2004}, apply error-correcting gradient descent learning only to the new synapses introduced between this new neuron and the hidden layer. In particular, if a novel property $m$ is ascribed to item $i$, we instantiate an additional element $\hat \y_m$ in the output layer of the neural network and add an additional row of weights $\wbr$ to $\wb$ representing new synaptic connections to this property neuron from the hidden layer. These weights are learned through gradient descent to attain the desired property value. Notably, we do not change other weights in the network (such as those from the input to the hidden layer), so as to prevent this fast learning of a single property from interfering with the broader bulk of knowledge already stored in the network (see \cite{McCloskey1989,McClelland1995} for a discussion of the catastrophic interference that arises from rapid non-interleaved learning). This yields the weight update
\begin{eqnarray*}
	 \tau_f\ddt \wbr & = & \frac{\partial}{\partial \wbr} \frac{1}{2} (\y_m^i - \hat \y_m^i)^2\\
	 		  	& = & (1-\wbr \h_i)\h_i^T
\end{eqnarray*}
where we have assumed the desired value for the feature $\y_m^i=1$ and $\h_i=\br\sqrt{\at}\rs^T\x^i$ is the hidden representation of item $i$. Here the time constant $\tau_f$ can be substantially faster than the time constant $\tau$ driving the development process. In this case, the dynamics above will converge rapidly to a steady state. If the new synaptic weights start at zero ($\wbr(0)=0$), they converge to 
\begin{equation*}
	\wbr =  \h_i^T/||\h_i||_2^2,
\end{equation*}
mirroring the hidden representation but with an appropriate rescaling. With these weights set, we may now ask how this knowledge will be extended to another item $j$ with hidden representation $\h_j$. The network's prediction is
\begin{eqnarray}
	\hat \y_m^j & = & (\wb\wa)_{mj}, \nonumber \\
			& = & \wbr \h_j \nonumber, \\
			& = & \h_i^T\h_j/||\h_i||_2^2, \nonumber
\end{eqnarray}
yielding Eqn.\newprop of the main text.
Hence generalization occurs in proportion to the overlap in hidden representations between the familiar item $i$ to which the new property $m$ was ascribed and the familiar probe item $j$.

A parallel situation exists for learning that a novel item $i$ possesses a familiar feature $m$. We add a new input node $\x_i$ to the network corresponding to the novel item. This input node is connected to the hidden layer through a new set of synaptic weights $\war$ which form a new column of $\wa$. To leave the knowledge in the network intact, we perform gradient learning on only these new connections, corresponding to the ``backpropagation to representation'' procedure used by \cite{Rogers2004}. Define $\h_m$ to be the transpose of the $m$th row of $\ls \sqrt{\at}\br^T$, that is, the ``backpropagated'' hidden representation of feature $m$. Then
\begin{eqnarray*}
	 \tau_f\ddt \war & = & \frac{\partial}{\partial \war} \frac{1}{2} (\y_m^i - \hat \y_m^i)^2\\
				 %& = & \frac{\partial}{\partial \war} \frac{1}{2} (\y_m^i - \wb\wa \x_i)^2\\
	 		  	& = & (1-\h_m^T \war)\h_m
\end{eqnarray*}
where we have assumed that the familiar feature has value $\y_m^i=1$ and the novel input $\x^i$ is a one-hot vector with its $i$th element equal to one and the rest zero. Solving for the steady state (starting from zero initial weights $\war(0)=0$) yields weights
\begin{equation*}
	\war =  \h_m/||\h_m||_2^2.
\end{equation*}
With these weights configured, the extent to which the novel object $i$ will be thought to have another familiar feature $n$ is
\begin{eqnarray}
	\hat \y_n^i & = & (\wb\wa)_{ni}, \nonumber \\
			& = & \h_n^T\war \nonumber, \\
			& = & \h_n^T\h_m/||\h_m||_2^2, \nonumber
\end{eqnarray}
yielding Eqn.\newitem of the main text.

\section{Linking Behavior and Neural Representations}

\subsection{Similarity structure is an invariant of optimal learning}
Here we show that two networks trained on the same statistics starting from small random initial conditions will have identical similarity structure in their hidden layer representations. This relation does not hold generally, however: hidden activity similarity structure can vary widely between networks that still perform the same input-output task. We show that identical similarity structure arises only in networks that optimally implement the desired task in the sense that they use the minimum norm weights necessary to implement the input-output mapping. 

The neural activity patterns in response to a set of probe items $\bx$, concatenated columnwise into the matrix $\bh$, is given by
\begin{eqnarray}
	\bh_1 & = & \br_1 \sqrt{\at}\rs^T \bx \nonumber\\
	\bh_2 & = & \br_2 \sqrt{\at}\rs^T\bx. \nonumber
\end{eqnarray}
Hence the similarity structure $\bh^T\bh$ is identical in both models, since
\begin{eqnarray}
	\bh_1^T\bh_1 & = &  \bx^T\rs \sqrt{\at} \br_1^T\br_1 \sqrt{\at}\rs^T\bx \nonumber\\
	 & = & \bx^T\rs \at \rs^T\bx \nonumber\\
	 & = &  \bx^T\rs \sqrt{\at} \br_2^T\br_2 \sqrt{\at}\rs^T\bx \nonumber\\
	 & = & \bh_2^T\bh_2. \nonumber
\end{eqnarray}
The key fact is simply that the arbitrary rotations are orthogonal, such that $\br_1^T\br_1=\br_2^T\br_2={\bf I}$. 

This invariance of the hidden similarity structure does not hold in general. Networks can perform the same input-output task but have widely different internal similarity structure. The full space of weight matrices that implement the desired input-output map is given by
\begin{eqnarray}
\wa(t) &  = & \bq \sqrt{\at}\rs^T, \label{eq:full_weight_degeneracy_a}\\
 \wb(t)  & = &  \ls  \sqrt{\at}\bq^{-1} \label{eq:full_weight_degeneracy_b}
\end{eqnarray}
That is, the ambiguity in neural representations arising from degeneracy in the solutions is given by any invertible matrix $\bq$. In this more general case, two networks will no longer have identical similarity structure since
\begin{eqnarray*}
	\bh_1^T\bh_1 & = &  \bx^T\rs \sqrt{\at} \bq_1^T\bq_1 \sqrt{\at}\rs^T\bx \\
	 & \neq &  \bx^T\rs \sqrt{\at} \bq_2^T\bq_2 \sqrt{\at}\rs^T\bx \nonumber\\
	 & = & \bh_2^T\bh_2, \nonumber
\end{eqnarray*}
because $\bq^T\bq \neq {\bf I}$. 

Why is the ambiguity in neural representations, encoded by the matrices $\br$, necessarily orthogonal in the learned solution from \textit{tabula rasa}? A well-known optimality principle governs this behavior: among all weight matrices that implement the desired input-output map, these solutions have minimum norm. We prove this here for completeness. 

Consider the problem
\begin{eqnarray*}
    \min_{\wb,\wa} || \wb ||_F^2 + ||\wa||_F^2 \\
    \textrm{s.t.} \quad \wb\wa=\ls \sv \rs^T
\end{eqnarray*}
in which we seek the minimum total Frobenius norm implementation of a particular input-output mapping. We can express the space of possible weight matrices as 

\begin{eqnarray*}
    \wa &  = & \bq \ba\rs^T, \label{eq:full_weight_degeneracy_a2}\\
    \wb  & = &  \ls \ba \bp \label{eq:full_weight_degeneracy_b2}
\end{eqnarray*}
where $\ba=\sqrt{\sv}$ and we enforce the constraint $\bp\bq={\bf I}$. This yields the equivalent problem
\begin{eqnarray*}
    \min_{\bp,\bq} || \wb ||_F^2 + ||\wa||_F^2 \\
    \textrm{s.t.} \quad \bp\bq={\bf I}.
\end{eqnarray*}
We will show that a minimizer of this problem must have $\bp=\br^T$ and $\bq=\br$ for some orthogonal matrix $\br$ such that $\br^T\br={\bf I}$. 

To solve this we introduce Lagrange multipliers ${\bf \Lambda}$ and form the Lagrangian
\begin{eqnarray*}
\mathcal{L}&=& || \wb ||_F^2 + ||\wa||_F^2 + \tr \left[{\bf \Lambda}^T(\bp\bq-{\bf I})\right] \\
    &=&\tr\left[\bp\bp^T\ba^2\right]+\tr\left[\bq^T\bq\ba^2 \right]\\
    &&+ \tr \left[{\bf \Lambda}^T(\bp\bq-{\bf I})\right].
\end{eqnarray*}
Differentiating and setting the result to zero we obtain
\begin{eqnarray*}
    \frac{\partial \mathcal{L}}{\partial \bp} & = & 2\ba^2\bp+{\bf \Lambda}\bq^T=0\\
    \frac{\partial \mathcal{L}}{\partial \bq} & = & 2\bq\ba^2+\bp^T{\bf \Lambda}=0\\
    \implies {\bf \Lambda} & = & -2\ba^2\bp\bq^{-T} = -2\bp^{-T}\bq\ba^2.
\end{eqnarray*}

Now note that since $\bp\bq={\bf I}$, we have $\bq=\bp^{-1}$ and $\bp^T=\bq^{-T}$, giving
\begin{eqnarray}
    -2\ba^2\bp\bq^{-T} &=& -2\bp^{-T}\bq\ba^2 \nonumber\\
 %   \ba^2\bp\bp^{T} &=& \bp^{-T}\bp^{-1}\ba^2 \nonumber\\
    \ba^2\bp\bp^{T} &=& (\bp\bp^{T})^{-1}\ba^2 \nonumber \\
    \sv \bm & = & \bm^{-1}\sv \label{eq:bm_constraint}
\end{eqnarray}
where we have defined $\bm \equiv \bp\bp^T$. Decomposing $\wb$ with the singular value decomposition,
\begin{eqnarray*}
    \wb&=&\ls \tilde \ba \tilde \rs^T = \ls \ba \bp\\
     \implies \bp &= & \ba^{-1}\tilde \ba \tilde \rs^T \\
                & = & \bd \tilde \rs^T
\end{eqnarray*}
where $\bd\equiv\ba^{-1}\tilde \ba$ is a diagonal matrix. Hence $\bm=\bp\bp^T=\bd^2$, so $\bm$ is also diagonal. Returning to Eqn.~\req{eq:bm_constraint}, we have
\begin{eqnarray*}
    \bm\sv & = & \bm^{-1} \sv\\
    \bm^2\sv& = &\sv
\end{eqnarray*}
where we have used the fact that diagonal matrices commute. To satisfy this expression, elements of $\bm$ on the diagonal must be $\pm1$ for any nonzero elements of $\sv$, but since $\bm=\bd^2$ we must select the positive solution. For elements of $\sv$ equal to zero, $\bm_{ii}=1$ still satisfies the equation (weights in these directions must be zero). This yields $\bm={\bf I}$, and so $\bp\bp^T={\bf I}$. Therefore $\bp$ is orthogonal. Finally $\bq=\bp^{-1}=\bp^T$, and so is orthogonal as well.  

Minimum norm implementations of a network's input-output map thus have the form
\begin{eqnarray*}
\wa(t) &  = & \br \sqrt{\at}\rs^T, \label{eq:full_weight_degeneracy_a3}\\
 \wb(t)  & = &  \ls  \sqrt{\at}\br^{T} \label{eq:full_weight_degeneracy_b3}
\end{eqnarray*}
where the ambiguity matrix $\br$ is orthogonal, $\br^T\br={\bf I}$. This is identical to the form of the weights found under $\textit{tabula rasa}$ learning dynamics, showing that gradient learning from small initial weights naturally finds the optimal norm solution.

\subsection{When the brain mirrors behavior}
The behavioral properties attributed to each item may be collected into the matrix $\by=\wb(t) \wa(t)\bx$. Its similarity structure $\by^T\by$ is thus
\begin{eqnarray}
	\by^T\by&=&\bx^T\wa(t)^T\wb(t)^T\wb(t)\wa(t) \bx \nonumber\\
	& = & \bx^T\rs \at\ls^T\ls \at \rs^T\bx \nonumber\\
	& = & \bx^T\rs \at^2 \rs^T \bx \nonumber\\
	& = & \left(\bh^T\bh\right)^2, \nonumber
\end{eqnarray}
where in the last step we have used the assumption that the probe inputs are white ($\bx^T \bx={\bf I}$), such that they have similar statistics to those seen during learning (recall $\si={\bf I}$ by assumption). This yields Eqn.\neuralrepbehav of the main text. We note that, again, this link between behavior and neural representations emerges only in optimal minimum norm implementations of the input-output map.

Hence the behavioral similarity of items shares the same object-analyzer vectors, and therefore the same categorical structure, as the neural representation; but each semantic distinction is expressed more strongly (according to the square of its singular value) in behavior relative to the neural representation. Intuitively, this greater distinction in behavior is due to the fact that half of the semantic relation is encoded in the output weights $\wb$, which do not influence the neural similarity of the hidden layer, as it depends only on $\wa$. 

\subsection{Simulation details for linking behavior and neural representations}
Here we describe the experimental parameters for \figlearninginvar~of the main text. We trained networks on a minimal hand-crafted hierarchical dataset with $N_3=7$ features, $N_2=32$ hidden units, and $N_1=P=4$ items. Inputs were encoded with one-hot vectors. The dataset was given by
\begin{eqnarray*}
    \sio &=& 0.7P\begin{bmatrix}
    1 & 1 & 1 & 1 \\
    1 & 1 & 0 & 0 \\
    0 & 0 & 1 & 1 \\
    1 & 0 & 0 & 0 \\
    0 & 1 & 0 & 0 \\
    0 & 0 & 1 & 0 \\
    0 & 0 & 0 & 1
    \end{bmatrix}\\
    \si &=& {\bf I}.
\end{eqnarray*}

The full-batch gradient descent dynamics were simulated for four networks with $\lambda=0.01$ for a thousand epochs. Networks were initialized with independent random Gaussian weights in both layers,
\begin{eqnarray*}
    \wa(0)_{ij} & \sim & \mathcal{N}(0,a_0^2/N_1) \\
     \wb(0)_{ij} & \sim & \mathcal{N}(0,a_0^2/N_3).
\end{eqnarray*}
The two small-initialization networks (panels A-B) had $a_0=0.0002$ while the two large initialization networks (panels C-D) had $a_0=1$. Individual neural responses and representational similarity matrices from the hidden layer and behavior were calculated at the end of learning, using probe inputs corresponding to the original inputs ($\bx={\bf I}$).

\bibliographystyle{unsrt}
\bibliography{semantic_cog_supp_mat}

\end{article}